\documentclass[journal]{IEEEtran} 
\usepackage{amsmath,amsfonts}
\usepackage{algorithmic}
\usepackage{algorithm}
\usepackage{array}
\usepackage[caption=false,font=footnotesize]{subfig}
\usepackage{textcomp}
\usepackage{stfloats}
\usepackage{url}
\usepackage{verbatim}
\usepackage{cite}
\usepackage{booktabs}
\usepackage[table]{xcolor}
\usepackage{makecell} 
\usepackage{multirow}

\usepackage{graphicx}
\usepackage{wrapfig}
\usepackage{float}
\usepackage{tikz}
\usetikzlibrary{patterns}
\usepackage[export]{adjustbox}
\usepackage[percent]{overpic}
\usetikzlibrary{calc}

\usepackage{pgfplots}
\pgfplotsset{compat=1.18}
\usepgfplotslibrary{groupplots}

\usepackage{hyperref}
\hypersetup{
    colorlinks=true,
    linkcolor=blue,
    filecolor=magenta,      
    urlcolor=cyan,
    pdftitle={Overleaf Example},
    pdfpagemode=FullScreen,
}

\definecolor{darkgreen}{RGB}{0,100,0}
\definecolor{myblue}{HTML}{5883DB}
\definecolor{mygreen}{HTML}{4E975F}

\definecolor{visionBlue}{HTML}{6CA6CD}   
\definecolor{tacRGBOrange}{HTML}{E8A868} 
\definecolor{tacFFRed}{HTML}{C76A6A}     

\begin{document}

\title{ManiFeel: Benchmarking and Understanding Visuotactile Manipulation Policy Learning}


\author{%
Quan Khanh Luu$^{*}$, Pokuang Zhou$^{*}$, Zhengtong Xu$^{*}$, Zhiyuan Zhang, Qiang Qiu, and Yu She%
\thanks{$^{*}$Equal contribution.}%
\thanks{All authors are with Purdue University, West Lafayette, IN, USA
(e-mail: luu15@purdue.edu; zhou1458@purdue.edu; xu1703@purdue.edu; zhan5570@purdue.edu; qqiu@purdue.edu; shey@purdue.edu).}%
}

\maketitle

\begin{abstract}
Supervised visuomotor policies have shown strong performance in robotic manipulation but often struggle in tasks with limited visual inputs, such as operations in confined spaces and dimly lit environments, or tasks requiring precise perception of object properties and environmental interactions.
In such cases, tactile feedback becomes essential for manipulation. While the rapid progress of supervised visuomotor policies has benefited greatly from high-quality, reproducible simulation benchmarks in visual imitation, the visuotactile domain still lacks a similarly comprehensive and reliable benchmark for large-scale and rigorous evaluation. To address this, we introduce ManiFeel, a reproducible and scalable simulation benchmark designed to systematically study supervised visuotactile policy learning.
ManiFeel offers a diverse suite of contact-rich and visually challenging manipulation tasks, a modular evaluation pipeline spanning sensing modalities, tactile representations, and policy architectures, as well as real-world validation.
Through extensive experiments, ManiFeel demonstrates how tactile sensing enhances policy performance across diverse manipulation scenarios, ranging from precise contact-driven operations to visually constrained settings. In addition, the results reveal task-dependent strengths of different tactile modalities and identify key design principles and open challenges for robust visuotactile policy learning. Real-world evaluations further confirm that ManiFeel provides a reliable and meaningful foundation for benchmarking and future visuotactile policy development.
To foster reproducibility and future research, we will release our codebase, datasets, training logs, and pretrained checkpoints, aiming to accelerate progress toward generalizable visuotactile policy learning and manipulation. 
\end{abstract}


\section{Introduction} \label{sec:introduction}
In recent years, supervised policy learning has made significant strides in robot manipulation, where visual input is used to generate actions for long-horizon and dexterous manipulation tasks~\cite{chi2023diffusionpolicy,chi2024universal,zhao2023learning,fu2024mobile}.
However, vision-based policies face notable limitations not only in environments where visual cues are absent or severely degraded, but also in manipulation scenarios that inherently demand precise contact interactions. In cluttered or confined spaces, under low-light conditions, or during complex contact-rich tasks, relying solely on vision becomes insufficient for robust policy execution. In such contexts, tactile feedback provides essential complementary information, capturing contact geometry, force distribution, and subtle interaction dynamics that vision alone cannot perceive.

Combining visual and tactile modalities has therefore become a promising and intuitive direction for learning robotic manipulation, particularly within supervised policy learning frameworks. Recent studies have shown that integrating high-resolution vision-based tactile sensors, such as GelSight~\cite{yuan2017gelsight}, can significantly enhance manipulation performance across diverse contact-rich and visually challenging scenarios~\cite{wang2024poco,yu2023mimictouch,xu2025unit,xue2025reactive,zhao2024transferable}. Other tactile sensing modalities, including tactile arrays~\cite{guzey2023see,lin2024learning,bhirangi2024anyskin}, force-torque sensors~\cite{yang2023seq2seq,hou2024adaptive}, tactile 3D point clouds~\cite{huang2024dvitac}, and contact-induced audio signals~\cite{liu2024maniwav}, have likewise demonstrated strong potential for enhancing perception and control in contact-intensive or unstructured environments.

However, unlike visual imitation learning, which benefits from a wide range of established simulation benchmarks that support large-scale experimentation and statistically meaningful comparisons \cite{mandlekar2022matters,mandlekar2023mimicgen,tao2024maniskill3}, visuotactile policy learning lacks comparable infrastructure. This gap is largely attributed to the challenges of accurately simulating high-fidelity tactile sensing. The lack of standardized benchmarking tasks and reproducible simulation environments has hindered the systematic development and evaluation of robust visuotactile manipulation policies. Consequently, evaluating the effectiveness of different approaches in a controlled and consistent manner remains a significant challenge. As a result, current research in supervised visuotactile policy learning lacks sufficient comparative analysis \cite{zhao2024transferable, xu2025unit, yu2023mimictouch}, limiting our understanding of its core principles and hindering the exploration of promising future directions.

To address these challenges, we introduce \textbf{ManiFeel}, a scalable and reproducible simulation benchmark for supervised visuotactile policy learning. ManiFeel provides a comprehensive testbed for analyzing the strengths and limitations of different sensing modalities, and offering insights into visuotactile policy design. Our contribution can be summarized as follows.

1) \textbf {Visuotactile Task Suite}: ManiFeel offers a diverse set of contact-rich manipulation tasks and corresponding human demonstration datasets, covering insertion, assembly, object exploration inside containers, and sorting tasks under visually challenging conditions. These tasks are designed to capture a wide range of contact-rich and visually limited scenarios, enabling systematic evaluation of how different tactile information contributes to multimodal policy learning. 

2) \textbf {Modular Policy Architecture Design}: ManiFeel modularizes the visuotactile policy pipeline into three key components: sensing modalities, tactile representations, and policies. This modular design facilitates the integration of diverse policy architectures. By enabling flexible combinations and systematic comparisons across these modules, ManiFeel enables in-depth investigation into which model structures offer greater advantages for visuotactile policy learning.

3) \textbf{Empirical Analysis of Visuotactile Policy Learning}: We conduct extensive empirical comparisons on diverse tasks across different policy architectures by systematically varying sensing modalities, tactile representations, and policies. These experiments are conducted in both simulation and real-world settings, enabling us to identify how different tactile representations support different types of tasks, uncover effective design principles for visuotactile policy learning, and highlight key challenges and promising directions for future research.

4) \textbf{Real-world Deployment and Reproduction}: In addition to constructing simulation benchmarks, we also reproduce the results in real-world settings, comparing different sensing modalities. The correlation between real-world and simulation results further validates the reliability and practical relevance of the ManiFeel simulation benchmark.

5) \textbf{Open-Source and Reproducible Benchmark Platform}:
To facilitate reproducible research and accelerate progress in visuotactile manipulation, ManiFeel is fully open-sourced, including its codebase, dataset, and pretrained checkpoints. The benchmark provides a standardized and easily extensible platform that enables fair comparisons and systematic evaluations across different models and sensing configurations. By establishing shared baselines and transparent experimental protocols, ManiFeel lowers the entry barrier for new research and promotes cumulative progress toward more generalizable and robust visuotactile manipulation in diverse real-world scenarios.

\begin{figure*}[t]
      \centering
  \includegraphics[width=0.80\linewidth]
  {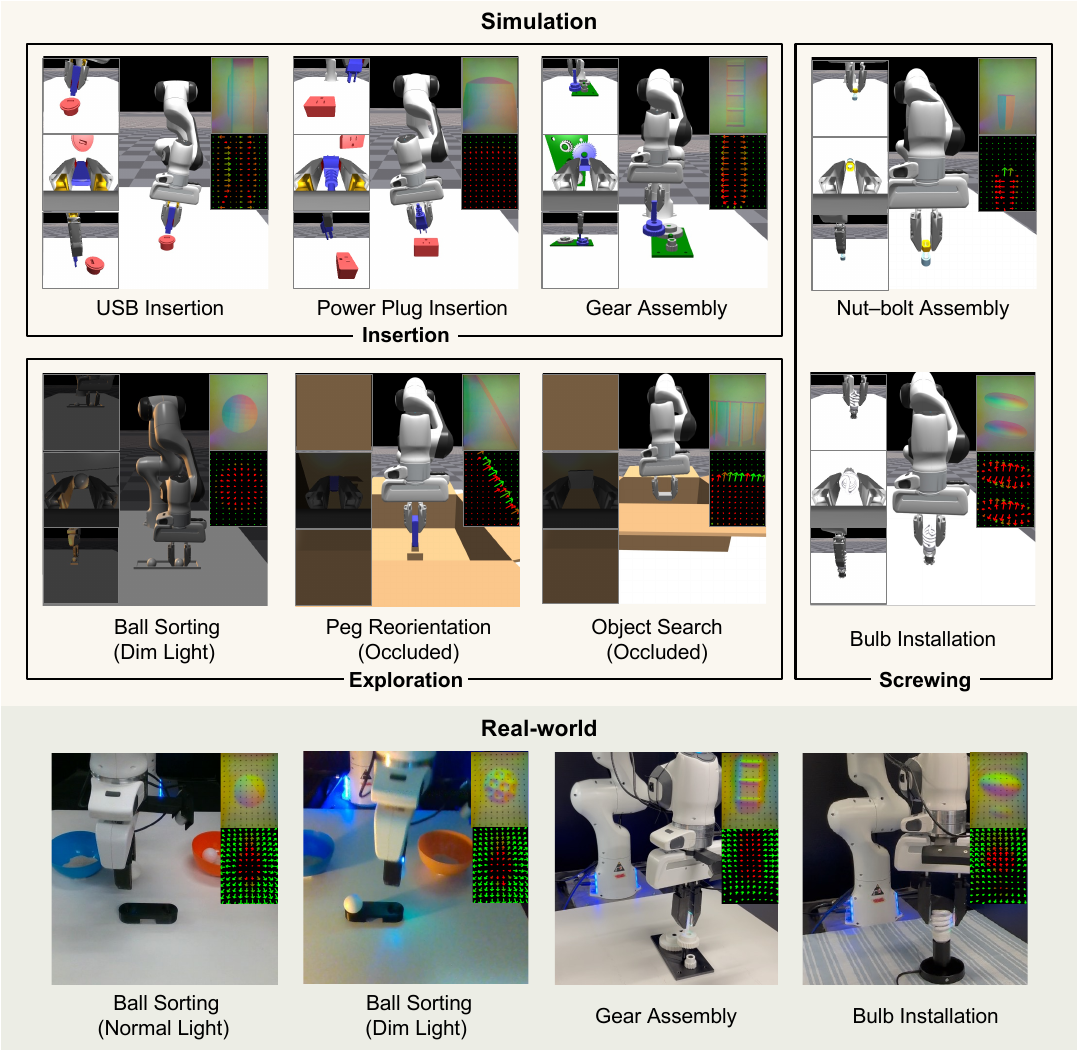}
    \caption{Overview of the manipulation tasks in the ManiFeel benchmark, showing each task's setup along with the available visual modalities (e.g., wrist camera, agentview camera) and tactile observations (e.g., high-resolution tactile images and tactile force-field signals).}
      \label{fig:manifeel_overview}
\end{figure*}
\section{Related Work}

\subsection{Imitation Learning}

In recent years, imitation learning has progressed rapidly, enabling robots to perform complex, dexterous, and long-horizon manipulation within supervised learning paradigm \cite{chi2023diffusionpolicy,chi2024universal,zhao2023learning,fu2024mobile}. Different lines of work have focused on various aspects of imitation learning, such as novel data collection pipelines \cite{zhao2023learning,fu2024mobile,wang2024dexcap}, the design of diverse policy heads such as diffusion \cite{chi2023diffusionpolicy}, flow matching \cite{zhang2024affordance, zhang2024flowpolicy}, and VQVAE \cite{lee2024behavior}, and the development of large-scale policies by integrating vision-language models \cite{black2410pi0} or scaling up both dataset and model capacity \cite{team2023octo,zhao2024aloha}.

Traditional imitation learning has predominantly relied on visual modalities, such as images \cite{chi2023diffusionpolicy,chi2024universal,zhao2023learning} or 3D point clouds \cite{ze20243d,xue2025demogen,huang2025vtrefine,zhu2025touch}. Moreover, visuotactile imitation learning has attracted growing attention. Since manipulation inherently involves interactions with the environment, tactile sensing provides complementary information that is crucial for capturing these interactions. As such, combining visual and tactile modalities to learn manipulation policies via imitation learning has become a natural and promising direction. For instance, several works have leveraged high-resolution vision-based tactile sensors such as GelSight \cite{yuan2017gelsight}, which provide rich contact information and have led to notable improvements in policy learning \cite{wang2024poco,yu2023mimictouch,xu2025unit,xue2025reactive}. In parallel, other tactile modalities, including tactile arrays \cite{guzey2023see,lin2024learning,bhirangi2024anyskin}, force-torque sensors \cite{yang2023seq2seq,hou2024adaptive}, tactile 3D point cloud \cite{huang2024dvitac}, and audio sensors \cite{liu2024maniwav}, have also demonstrated strong performance, particularly in tasks involving fine-grained manipulation and complex environment interactions.

However, unlike visual imitation learning, which benefits from a wide range of established policy learning benchmarks, tactile sensing remains difficult to simulate accurately in existing robotic simulators. As a result, most prior works report results directly in real-world environments, without simulation-based benchmarks. Yet, simulation benchmarks offer significant advantages: they allow for extensive rollouts across diverse environments and enable consistent evaluation throughout training, leading to more statistically meaningful comparisons and insights into learning trends. This is particularly valuable for understanding the role of tactile feedback in imitation learning.

Despite this potential, there is currently a lack of simulation benchmarks tailored to visuotactile policy learning. To address this gap, we introduce a new benchmark specifically designed for evaluating supervised visuotactile policy learning in simulated settings.

\subsection{Visuotactile Manipulation}

Tactile feedback has been widely explored for manipulation tasks involving rigid \cite{hogan2020tactile,kim2021active,shirai2023tactile,du2024stick}, deformable \cite{she2021cable,sunil2022visuotactile,10947008,zhou2025hand}, and articulated objects \cite{zhao2024tac}. A common strategy is to extract low-level state estimations, such as contact areas or object pose, from tactile signals to guide model-based controllers. Moreover, some approaches integrate tactile signals into model predictive control frameworks \cite{tian2019manipulation, quan2025protac}, or into task-specific solutions like page flipping \cite{zheng2022autonomous}, grasping \cite{xu2024letac}, robotic pushing \cite{lloyd2021goal, quan2023simtacls}. Finally, learning-based tactile control frameworks have emerged to improve adaptability. For instance, neural networks have been used to predict grasp stability or success \cite{calandra2017feeling,si2022grasp,kanitkar2022poseit}, and to evaluate candidate actions by sampling-based method \cite{calandra2018more,hogan2018tactile,feng2020center,han2021learning}.

The aforementioned works focus on modeling specific tasks or object categories by incorporating visuotactile information, and subsequently designing manipulation policies. However, the resulting policies or controllers are typically task-specific, and their pipelines often struggle to generalize to other tasks or object types. In contrast, imitation learning offers a more general and flexible approach by learning manipulation policies directly from demonstration data, making it more adaptable to a broader range of tasks and objects.
 
\subsection{Tactile Representation Learning}
Recent advances in tactile representation learning span a wide range of approaches. Some studies leverage convolutional neural networks to extract features directly from tactile signals \cite{polic2019convolutional}, while others use masked autoencoding methods for more structured representation learning \cite{sferrazza2023power, cao2023learn}. The work in \cite{xu2025unit} employs VQGAN to learn a generalizable efficient tactile representation using very simple object dataset. In parallel, growing efforts have focused on developing transferable representations across diverse tactile sensors and perception tasks \cite{higuera2024sparsh, zhao2024transferable}. The study in \cite{gupta2025sensor} introduces a sensor-invariant tactile representation by incorporating a normal map reconstruction loss. Notably, \cite{yang2024unitouch, feng2025anytouch} propose multimodal representations by grounding tactile signals in pretrained foundation models, enabling improved generalization across modalities.

Most of the existing tactile representation learning works evaluate their methods on downstream perception tasks such as classification and pose estimation, as these benchmarks are relatively easy to set up and require low data collection costs. In contrast, a fair and comprehensive comparison of different tactile representations in the context of policy learning is equally important but less explored. Our benchmark enables such evaluation by providing a standardized and scalable framework for policy learning with tactile representations.
\section{ManiFeel: Visuotactile Manipulation Learning Benchmark} \label{sec:manifeel}
This section introduces the task suite and modular pipeline design of the~ManiFeel benchmark. ManiFeel features a diverse set of manipulation tasks spanning clear-vision conditions to severely degraded or fully occluded visual settings (Section~\ref{sec:task_suite}). The benchmark includes three main task categories: \textit{Insertion}, \textit{Screwing}, and \textit{Exploration}. This results in a total of $13$ different task setups, including $9$ simulation setups and $4$ real-world setups, as shown in Fig.~\ref{fig:manifeel_overview}. These tasks are designed to vary in their reliance on visual and tactile modalities, providing a comprehensive framework for analyzing policy performance across different sensory challenges. In addition, we describe the modular pipeline design that enables systematic evaluation of different sensing modalities, tactile representations, and policies across the task suite (Section~\ref{sec:pipeline_design}).

\subsection{Task Suite and Dataset} \label{sec:task_suite}
\subsubsection{Task and Environment Design}
We design a suite of diverse manipulation tasks to systematically evaluate supervised visuotactile policy learning under varying sensory modalities and environmental conditions. The tasks are organized into three main categories: \textit{Insertion}, \textit{Screwing}, and \textit{Exploration} manipulation tasks (see Fig.~\ref{fig:manifeel_overview}). \textit{Insertion} consists of classical assembly tasks, ranging from peg-in-hole insertion to gear assembly. \textit{Screwing} includes nut-and-bolt threading and light-bulb installation. \textit{Exploration} involves tasks where visual input is occluded or limited, such as object reorientation and object search within a box, as well as object sorting under dim lighting, requiring the robot to rely primarily on tactile feedback.
All task environments are implemented in the IsaacGym simulator~\cite{IsaacGym}, and one representative task from each category is replicated in the real-world setup. This section provides an overview of each task category, while detailed configurations and environment parameters are described in Appendix~\ref{app:task_details}.

\textit{a) Insertion Tasks.} The insertion group represents a fundamental class of contact-rich manipulation problems in robotic assembly and insertion processes, which require precise alignment and constrained motion between objects. In ManiFeel, this category includes the following representative tasks:

\textbf{Peg and Plug Insertion (Simulation).}
The insertion group includes tasks that represent fundamental challenges in contact-rich manipulation, including peg-in-hole, USB plug, and power plug insertions. The policies are trained end-to-end from visuotactile demonstrations to learn coordinated motion and fine alignment behaviors under variations in the pose and geometry of the object–socket pairs (see Fig.~\ref{fig:sim_demo_trajectory_insertion}).

\textbf{Gear Assembly (Simulation and Real).}
The gear assembly task simulates an industrial scenario where the robot mounts a medium-sized gear onto a central shaft while ensuring proper meshing with side gears. The policies are trained end-to-end from visuotactile demonstrations to learn accurate alignment and insertion behaviors. The same task is reproduced in the real world to examine the correspondence between simulated and real visuotactile learning behaviors (see Fig.~\ref{fig:real_gear_process}).

\textit{b) Screwing Tasks.} The screwing group captures another important class of contact-rich manipulation problems that involve rotational motion and torque-sensitive interactions. These tasks require the policy to regulate contact forces and rotation during threaded or twist-based assembly. In ManiFeel, this category includes the following representative tasks:

\textbf{Nut\mbox{--}Bolt Assembly (Simulation).}
In this task, the robot inserts and screws a nut onto a bolt. The policy is trained end-to-end from visuotactile demonstrations to learn alignment and threading behaviors directly from sensory observations. This task requires the robot to regulate sufficient grasp force to prevent finger–object slippage and to ensure smooth rotation throughout the assembly process.

\textbf{Bulb Installation (Simulation and Real).}
The bulb installation task follows a similar setup, where the robot inserts and screws a light bulb into its socket. In this case, the threading interface is occluded from view, posing a challenge for precise insertion and tightening. The policy is trained end-to-end from visuotactile demonstrations to learn contact-guided rotational motion and tightening behaviors. In this task, maintaining sufficient grasp force during rotation and detecting tightening completion are essential for stable contact and successful task execution. The task is also implemented in the real world to assess simulation fidelity and compare visuotactile behaviors between simulated and physical environments (see Fig.~\ref{fig:real_bulb_process}).

\textit{c) Exploration Tasks.} The exploration group focuses on manipulation scenarios where visual information is limited or unavailable, requiring the robot to actively explore and infer object states through tactile interaction. These tasks highlight the importance of tactile sensing for object localization, reorientation, and sorting in occluded or low-light environments. This category in ManiFeel includes the following representative tasks:

\textbf{Peg Reorientation (Simulation).}
This task emulates how humans manipulate or orient objects in confined spaces without visual access. The robot begins by holding a square peg at an arbitrary orientation and must first perform in-hand self-alignment, then insert the peg into the socket once the correct orientation is achieved. The policy is trained end-to-end from visuotactile demonstrations to learn contact-guided in-hand rotation and insertion behaviors. During execution, the robot leverages external contact with the rim of the socket opening to perform controlled in-hand rotation and achieve vertical alignment. The external front-facing camera cannot observe the interior, simulating a vision-occluded environment and highlighting the role of tactile sensing in enabling precise in-hand manipulation (see Fig.~\ref{fig:peg_reorientation_process}).

\textbf{Object Search (Simulation).}
This task emulates how humans retrieve objects from a bag without visual access. The robot must locate, grasp, and lift a cube-shaped object randomly placed inside a closed container. The policy is trained end-to-end from visuotactile demonstrations to learn exploratory behaviors for localizing and manipulating unseen objects under fully occluded visual conditions. The task requires the robot to use tactile feedback to explore the container, detect contact events, and regulate its grasping actions, demonstrating the emergence of active tactile exploration in the absence of visual cues (see Fig.~\ref{fig:object_search_process}).

\textbf{Ball Sorting (Simulation and Real).}
This task emulates how humans distinguish and handle objects with similar appearance by relying on touch when visual cues are limited. The task involves manipulating two visually similar spherical objects, a table-tennis ball and a golf ball, which require the robot to perceive and respond to their distinct surface textures during sorting. In simulation, the policy is trained end-to-end from visuotactile demonstrations to localize, identify, and lift the designated golf ball. The same task is replicated in the real world to assess simulation fidelity and examine the correspondence between simulated and real visuotactile behaviors. In the physical setup, the robot must identify each ball and place it into its designated bowl. The task is implemented under both normal and dim lighting conditions to examine how tactile sensing complements vision by enabling robust object identification and manipulation when visual information is degraded or ambiguous (see Figs.~\ref{fig:ball_sorting_process} and~\ref{fig:real_sorting_process_dim}).

\subsubsection{Observation modalities}
\begin{figure}[t]
  \centering
  \includegraphics[width=0.98\linewidth]{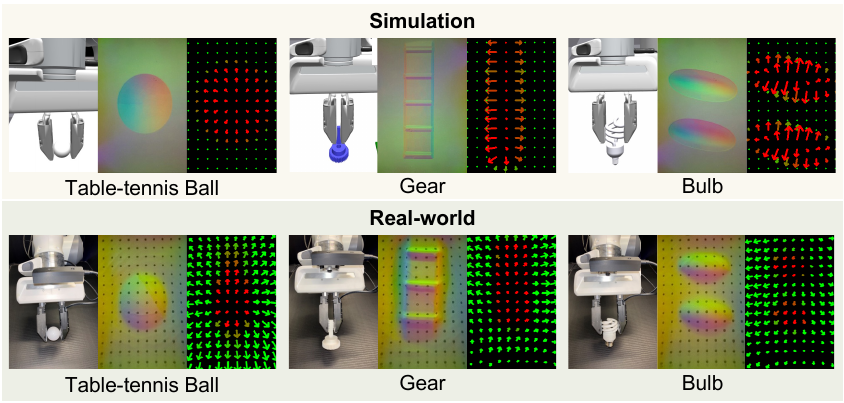}
  \caption{
  Comparison of simulated tactile data generated using TacSL \cite{akinola2024tacsl} and real tactile measurements captured from GelSight R1.5~\cite{wang2021gelsight}, including tactile RGB and tactile force-field signals.
We present examples featuring several objects included in the ManiFeel benchmark. }
  \label{fig:sim_real}
\end{figure}

ManiFeel is developed on the IsaacGym platform~\cite{IsaacGym} with the integration of the TacSL framework~\cite{akinola2024tacsl} to simulate robotic manipulators equipped with tactile fingers. 
TacSL supports high-fidelity simulation of vision-based tactile sensors based on the GelSight R1.5 design~\cite{wang2021gelsight}. 
The tactile modality includes two complementary observation channels. 
The first, tactile RGB images (TacRGB), are high-resolution tactile images that capture surface deformation patterns on the elastomer contact area, providing rich information about the contacted object's surface texture and geometry. 
The second, tactile force field (TacFF), represents spatial force-field data that describe distributed normal and shear forces across the contact surface, encoding both contact pressure and directional force information. 
Together, these tactile representations enable fine-grained perception of contact geometry and interaction dynamics during manipulation. 
Samples of simulated tactile observations and their close resemblance to real-world counterparts are shown in Fig.~\ref{fig:sim_real}, demonstrating the realism and potential for generalization of ManiFeel’s simulation benchmark. 
Samples of simulated tactile observations and their close resemblance to real-world counterparts are shown in Fig.~\ref{fig:sim_real}. To further validate this similarity, we quantitatively evaluate how well simulated tactile signals match real sensor measurements using two widely adopted distributional metrics: Fréchet Inception Distance (FID)~\cite{heusel2017fid} and Kernel Inception Distance (KID)~\cite{binkowski2018kid}. Both metrics operate on high-level visual features extracted by a ResNet encoder and measure the distance between the distributions of simulated and real tactile images, where lower values indicate stronger similarity. As summarized in Table~\ref{tab:sim_real_similarity}, both TacRGB and TacFF achieve consistently low FID and KID scores across representative contacted object instances (ball, bulb, gear).
The reported FID and KID values (below $7$ and $8\times10^{-2}$, respectively) fall well within the range generally considered indicative of close cross-domain alignment~\cite{heusel2017fid, zhu2017cyclegan}, confirming that simulated tactile feedback closely reproduces the statistical properties of real tactile observations. These quantitative results, together with qualitative visual comparisons, validate the realism of the simulated tactile feedback and support the role of ManiFeel as a reliable simulation benchmark for visuotactile policy learning.

For vision modality, the framework supports multiple camera configurations, such as front, side, or wrist-mounted views, and allows for occluded or degraded conditions to emulate realistic manipulation challenges. 
Detailed configurations of camera setups, lighting conditions, and observation modalities for each task are provided in Appendix~\ref{app:experiment_details}.

\begin{table}[t]
\centering
\caption{Sim--real tactile similarity evaluated using FID (Fréchet Inception Distance) and KID (Kernel Inception Distance). Lower scores indicate higher similarity.}
\label{tab:sim_real_similarity}
\vspace{2mm}
\begin{tabular}{lccc}
\toprule
\textbf{Modality} & \textbf{Object} & \textbf{FID} $\downarrow$ & \textbf{KID} ($\times 10^{-2}$) $\downarrow$ \\
\midrule
\multirow{3}{*}{\makecell[l]{\textbf{Tactile Image}\\(TacRGB)}} 
& Ball & 5.67 & 5.79 \\
& Bulb & 5.80 & 6.46 \\
& Gear & 6.57 & 7.44 \\
\midrule
\multirow{3}{*}{\makecell[l]{\textbf{Tactile Force-field}\\(TacFF)}} 
& Ball & 2.96 & 2.77 \\
& Bulb & 3.54 & 3.23 \\
& Gear & 2.91 & 2.54 \\
\bottomrule
\end{tabular}
\end{table}

\subsubsection{Teleoperated Demonstration Data Collection}
We collect human demonstrations for supervised policy learning using a 3Dconnexion SpaceMouse\footnote{\url{https://3dconnexion.com/us/product/spacemouse-compact/}}. Tactile feedback plays a critical role during data collection, especially for tasks requiring precise contact reasoning or operating under visual occlusion and degraded lighting conditions. In these settings, operators rely heavily on real-time tactile observations displayed on a monitor to perceive subtle contact events for effective object manipulation, classification, and exploration. A summary of the collected demonstration data across all tasks is provided in Appendix~\ref{app:dataset_details}.

\begin{figure}[t]
  \centering

  \subfloat[USB insertion\label{fig:usb_process}]    
    {%
        \includegraphics[width=0.98\linewidth]{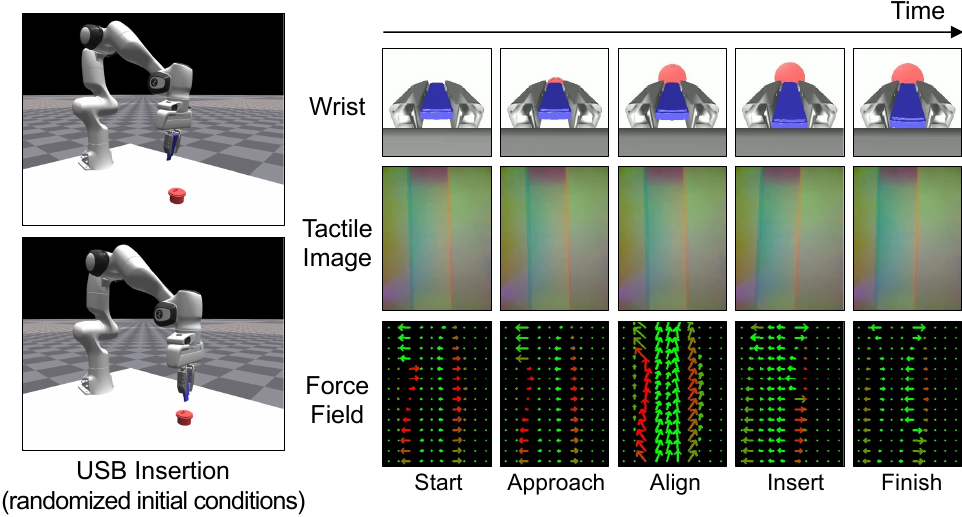}}
    
  \subfloat[Power plug insertion\label{fig:power_plug_process}]    
    {%
        \includegraphics[width=0.98\linewidth]{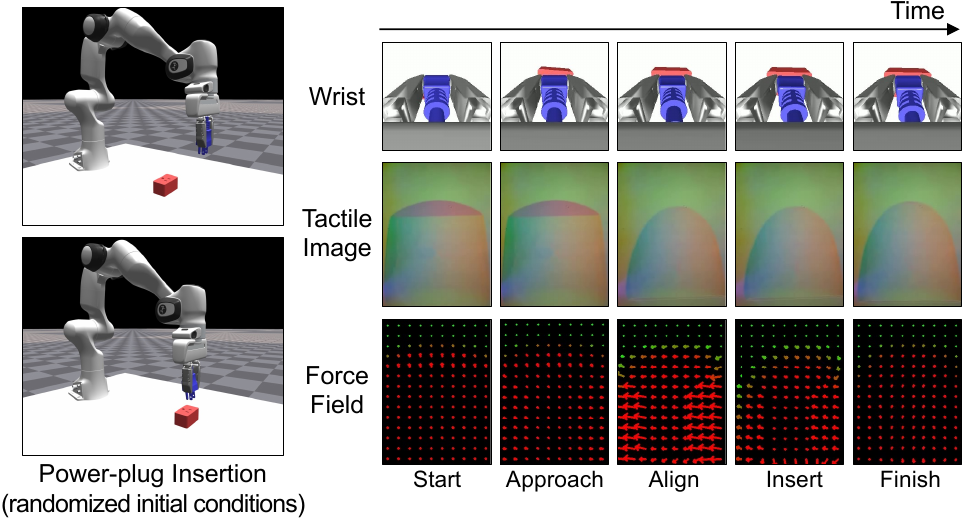}}
  
  \caption{Samples of teleoperated demonstration data in simulation for two representative insertion tasks: (a) USB insertion and (b) power plug insertion, highlighting contact-rich scenarios that demand precise spatial contact reasoning.}
  \label{fig:sim_demo_trajectory_insertion}
\end{figure}

\begin{figure}[t]
  \centering
  \subfloat[Peg reorientation\label{fig:peg_reorientation_process}]{
    \includegraphics[width=0.98\linewidth]{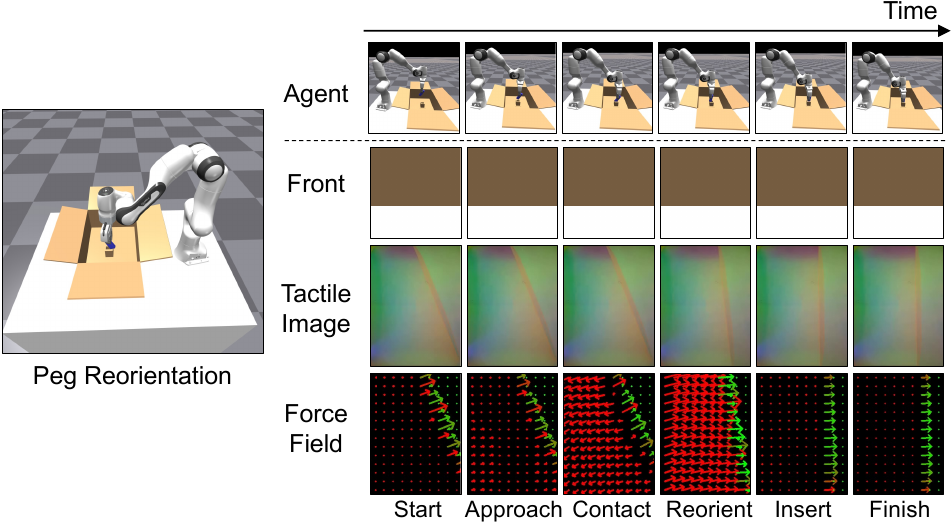}}
    \vspace{-2mm}
    
  \subfloat[Object search\label{fig:object_search_process}]{
    \includegraphics[width=0.98\linewidth]{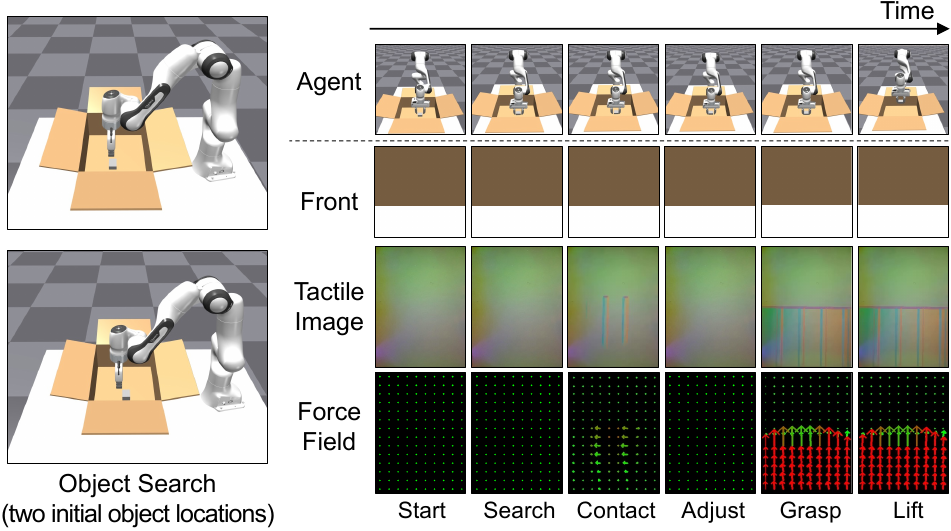}}
    \vspace{-2mm}

  \subfloat[Ball sorting\label{fig:ball_sorting_process}]{
    \includegraphics[width=0.98\linewidth]{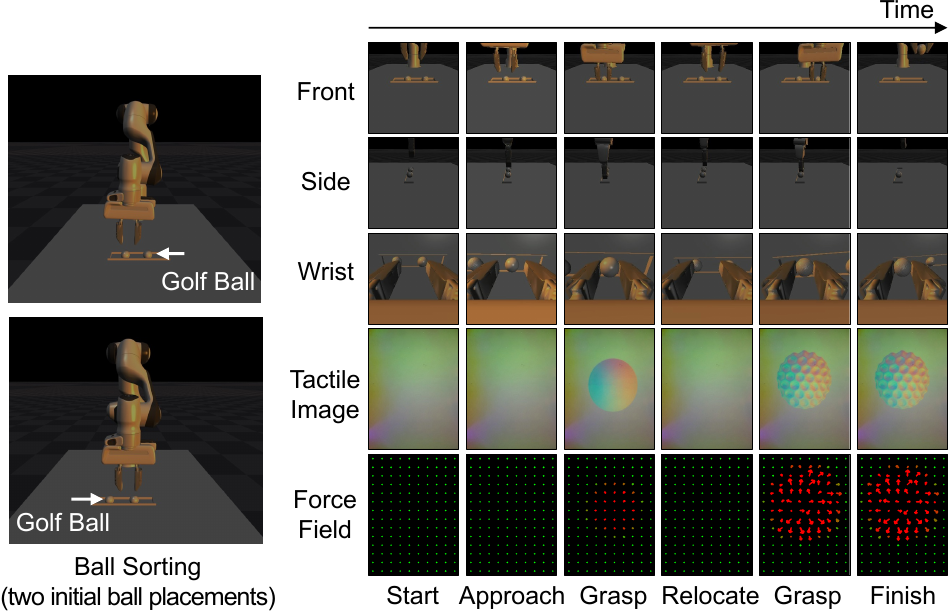}}
  
  \caption{Samples of teleoperated demonstration data in simulation for exploration tasks: (a) peg reorientation, (b) object search, and (c) ball sorting, highlighting scenarios with severely degraded or occluded visual input. The agent view is shown only for visualization of occluded scenes and is not included in the demonstration dataset.}
  \label{fig:sim_demo_trajectory}
\end{figure}

Representative examples of the teleoperated demonstration trajectories are visualized in Figs.~\ref{fig:sim_demo_trajectory_insertion} and \ref{fig:sim_demo_trajectory}. These examples illustrate how operators interpret multimodal feedback to guide fine-grained behaviors. For instance, in the tactile sorting task under dim lighting, visually similar objects such as a table tennis ball and a golf ball become difficult to distinguish from vision alone. Operators rely on tactile images to recognize differences in surface texture and to correct initial inaccurate grasps. Similar benefits arise in contact-rich insertion tasks, where tactile cues help estimate in-hand pose and detect misalignment that is not visible from external camera views. Such observations highlight that tactile feedback is essential not only for robust teleoperation but also for collecting high-quality demonstrations that support effective supervised visuotactile policy learning. All collected datasets will be open-sourced to support future research.

\subsection{Modular Pipeline Design} \label{sec:pipeline_design}
In this section, we present the modular pipeline design that forms the basis of our benchmark.
The benchmark is designed to systematically evaluate the performance of visuotactile policies which are composed of different sensing modalities, tactile representations, and policies, across a broad set of tasks. This modular structure enables flexible combinations and fair comparisons, ultimately facilitating a deeper understanding of the design principles behind effective visuotactile policy learning.
Moreover, the modularity of our design provides strong flexibility for integrating future advancements, such as new sensing modalities, novel tactile representation learning methods, and emerging policy architectures.

\textbf{Choice of Modalities}. 
To investigate the contribution of different sensing modalities, ManiFeel enables direct comparison across three sensing configurations: \textbf{vision-only}, \textbf{vision with TacRGB}, and \textbf{vision with TacFF}.
ManiFeel evaluates pairwise combinations of vision with a single tactile modality (either TacRGB or TacFF) to ensure clear and interpretable comparisons. By isolating each tactile modality, we can analyze how TacRGB and TacFF individually complement visual perception across different tasks and scenarios. Combining vision with both tactile modalities would introduce confounding interactions, making it difficult to attribute performance gains to a specific tactile representation and to understand its task-specific contribution. Additionally, we do not consider a tactile-only baseline because most benchmark tasks require visual information for global localization and pre-contact alignment, and our focus is to understand how tactile sensing augments a vision-based policy rather than replaces it.



\begin{figure}[t]
    \centering

    \subfloat[Vision + TacRGB observation encoder.]{%
    \label{fig:fusion_vistac}
        \includegraphics[width=0.98\linewidth]{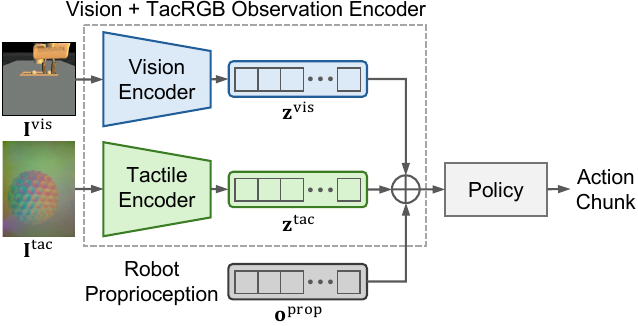}}

    \subfloat[Vision + TacFF observation encoder.]{%
    \label{fig:fusion_visff}
        \includegraphics[width=0.98\linewidth]{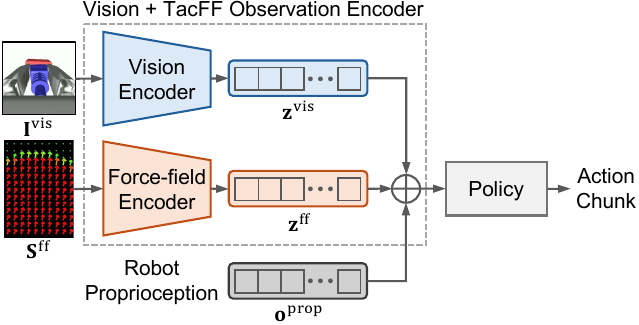}}

    \caption{Multimodal observation encoder architectures used in ManiFeel. Each configuration fuses visual observations ($\mathbf{I}^{\mathrm{vis}}$), tactile feedback ($\mathbf{I}^{\mathrm{tac}}$ or $\mathbf{S}^{\mathrm{ff}}$), and robot proprioception ($\mathbf{o}^{\mathrm{prop}}$) through separate modality encoders. The encoded features ($\mathbf{z}^{\mathrm{vis}}$, $\mathbf{z}^{\mathrm{tac}}$, and $\mathbf{z}^{\mathrm{ff}}$) are concatenated and passed to the policy network to generate action chunks.}
    \label{fig:fusion_strategies}
\end{figure}

Regarding the visuotactile sensing fusion scheme, ManiFeel employs a multimodal observation encoder architecture that integrates visual, tactile, and proprioceptive inputs, as illustrated in Fig.~\ref{fig:fusion_strategies}. Formally, at each timestep $t$, the multimodal observation includes both the current and previous observations from each modality, expressed as
\begin{equation}
\mathbf{o}_t = \{ \mathbf{I}^{\text{vis}}_{t-1:t},\ \mathbf{I}^{\text{tac}}_{t-1:t},\ \mathbf{S}^{\text{ff}}_{t-1:t},\ \mathbf{o}^{\text{prop}}_{t-1:t} \},
\end{equation}
where $\mathbf{I}^{\text{vis}} \in \mathbb{R}^{H \times W \times 3}$ denotes the visual observation from an RGB camera,
$\mathbf{I}^{\text{tac}} \in \mathbb{R}^{h \times w \times 3}$ represents the high-resolution tactile image (TacRGB),
$\mathbf{S}^{\text{ff}} \in \mathbb{R}^{r \times c \times 3}$ corresponds to the tactile force-field map (TacFF), encoding normal force and two-dimensional shear forces at each discretized contact point,
and $\mathbf{o}^{\text{prop}} \in \mathbb{R}^{7}$ denotes robot proprioception, consisting of the end-effector position and orientation represented in quaternion form.
All modalities are normalized to the range $[-1, 1]$ before being passed to their respective encoders.
Each sensing modality is processed by a dedicated encoder: $f_{\text{vis}}$ for visual observations, $f_{\text{tac}}$ for tactile RGB, and $f_{\text{ff}}$ for force-field tactile feedback. At each time step $t$, the encoder receives both the current and previous observations, denoted as $(\cdot)_{t-1:t}$, and computes the corresponding latent embeddings:
$\mathbf{z}^{\text{vis}}_t = f_{\text{vis}}(\mathbf{I}^{\text{vis}}_{t-1:t})\in \mathbb{R}^{d_{\text{vis}}}, \,
\mathbf{z}^{\text{tac}}_t = f_{\text{tac}}(\mathbf{I}^{\text{tac}}_{t-1:t})\in \mathbb{R}^{d_{\text{tac}}}, \,
\mathbf{z}^{\text{ff}}_t = f_{\text{ff}}(\mathbf{S}^{\text{ff}}_{t-1:t})\in \mathbb{R}^{d_{\text{ff}}}$.

For a given modality configuration (e.g., vision with TacRGB), the encoded latent embeddings are concatenated to form the fused multimodal representation:
\begin{equation}
\mathbf{z}_t = \mathbf{z}^{\text{vis}}_t \oplus \mathbf{z}^{\text{tac}}_t \oplus \mathbf{o}^{\text{prop}}_t \in \mathbb{R}^{d_{\text{vis}}+d_{\text{tac}}+7},
\end{equation}
or, for the TacFF configuration,
\begin{equation}
\mathbf{z}_t = \mathbf{z}^{\text{vis}}_t \oplus \mathbf{z}^{\text{ff}}_t \oplus \mathbf{o}^{\text{prop}}_t \in \mathbb{R}^{d_{\text{vis}}+d_{\text{ff}}+7},
\end{equation}
where $\oplus$ denotes feature concatenation. The fused representation $\mathbf{z}_t$ is then passed to the policy network to generate the action chunk.

\textbf{Tactile Representations}. While the choice of modality defines the type of tactile input (e.g., TacRGB or TacFF), ManiFeel also supports the integration and comparison of different tactile encoders ($f_{\text{tac}}$) that extract latent embeddings ($\mathbf{z}^{\text{tac}}$) from high-resolution tactile images ($\mathbf{I}^{\text{tac}}$). In this study, we benchmark four representative tactile representations: (1) ResNet18 trained from scratch \cite{chi2023diffusionpolicy,he2016deep}, (2) T3 \cite{zhao2024transferable}, a pretrained encoder using large-scale tactile datasets, (3) UniT \cite{xu2025unit}, a lightweight representation trained with generalizability, and (4) AnyTouch \cite{feng2025anytouch}, a generalizable tactile representation learned across diverse sensors and modalities. These comparisons aim to reveal whether pretrained or carefully designed tactile encoders offer meaningful advantages in visuotactile policy learning.

\textbf{Policies}. ManiFeel also supports benchmarking different policy architectures for visuotactile manipulation. In this study, we compare three representative policy types commonly used in robot learning and analyze their integration with different choices of visuotactile inputs: Diffusion Policy \cite{chi2023diffusionpolicy}, Equivariant Diffusion Policy \cite{wang2024equivariant}, and Flow Matching \cite{zhang2024flowpolicy}. These policies differ in their generative modeling approaches and model architectures. Our goal is to understand how each type of policy performs under different input modalities, and whether certain policies are more compatible with the inclusion of tactile sensing. Specifically, we study how tactile signals influence performance across policy architectures. Additional policy and pipeline details can be found in Appendix~\ref{app:pipeline_details}.
\section{Experiments and Results}
\label{sec:result}
We conduct a comprehensive set of experiments to evaluate the effectiveness of diverse visuotactile policies across diverse manipulation tasks using the ManiFeel simulation benchmark and real-world settings. Our study systematically benchmarks different sensory modalities, tactile representations, and policies under varying task conditions within the supervised policy learning framework. Note that all evaluated sensing configurations also include proprioceptive feedback (i.e., the robot end-effector's pose). For simulation results, success rates are reported as the primary evaluation metric across all experiments. Specifically, success rates are averaged over the last ten epochs of training, across 50 different environment initializations, and over three random seeds. This results in each reported success rate being the average of $1500$ evaluations, providing strong statistical significance. For real-world experiments, we perform $15$ consecutive rollouts and record the success rate. The initial environment conditions are varied for each rollout.
Through extensive experiments, we aim to address four central questions in supervised visuotactile policy learning:
\begin{enumerate}
    \item How does tactile sensing enhance performance in contact-rich and visually constrained manipulation tasks?
    \item How do different tactile modalities contribute to policy performance across diverse tasks?
    \item How consistent are the observed behaviors and outcomes between simulation and real-world environments?
    \item Lessons learned and design principles for developing efficient and robust visuotactile policies
\end{enumerate}
These questions guide our analysis toward understanding the role of tactile sensing in multimodal policy learning and identifying key challenges and future directions for advancing multimodal robotic manipulation.

\subsection{Modality Benchmarking}
\label{sec:modality-benchmark}
\textbf{Setup}. 
This section evaluates policy performance under three sensing configurations, vision only, vision with TacRGB, and vision with TacFF, across nine manipulation tasks, including four insertion tasks, two screwing tasks, and three exploration tasks, as defined in the ManiFeel simulation benchmark (see Section~\ref{sec:task_suite}).
In this setup, wrist-mounted camera views are used for the insertion and screwing task categories, while external occluded camera views are employed for the exploration tasks, particularly for peg reorientation and object search. The ball sorting task is conducted under dim lighting conditions to simulate visually degraded perception. This setup emulates the limited visual information available to humans when operating in similarly vision-occluded or low-light environments and highlights the challenges faced by visual perception in such scenarios.
All policies use a ResNet18 tactile encoder trained from scratch, and the policy is based on the Diffusion Policy \cite{chi2023diffusionpolicy}, a widely adopted and well-established framework for supervised policy learning. 
Further details regarding training details and observation configurations are provided in Appendix~\ref{app:experiment_details}.

\begin{figure}[t]
    \centering
      \includegraphics[width=\linewidth]{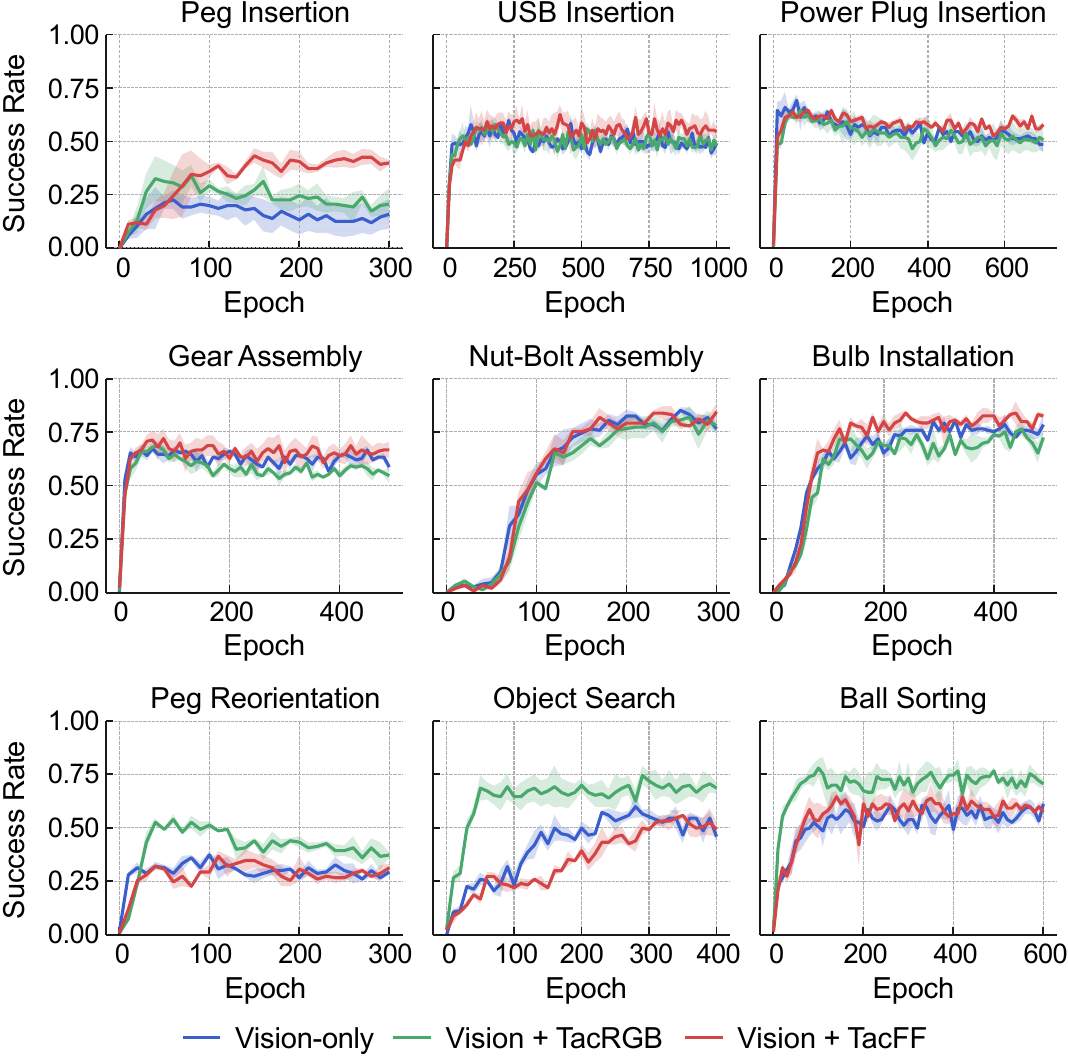}
    \caption{Learning curves comparing the three different sensing configurations (i.e., Vision-only, Vision + TacRGB, and Vision + TacFF) across $9$ simulation manipulation tasks. Each plot shows success rates over training epochs. Note that all sensing configurations also include robot proprioception inputs.}
    \label{fig:training_curves_modality}
\end{figure}

\begin{table*}[t]
\centering
\caption{Success rates (mean $\pm$ std) averaged over three seeds across tasks and sensing configurations. Best in each column is bold.}
\label{tab:modality_bechmarking}
\resizebox{\textwidth}{!}{%
\begin{tabular}{l|cccc|cc|ccc}
\toprule
\textbf{\makecell[l]{Sensing\\Configuration}} &
\multicolumn{4}{c|}{\textbf{Insertion}} &
\multicolumn{2}{c|}{\textbf{Screwing}} &
\multicolumn{3}{c}{\textbf{Exploration}} \\
\cmidrule(lr){2-5} \cmidrule(lr){6-7} \cmidrule(lr){8-10}
 & \makecell{Peg\\Insertion} &
   \makecell{USB\\Insertion} &
   \makecell{Power Plug\\Insertion} &
   \makecell{Gear\\Assembly} &
   \makecell{Nut\mbox{--}bolt\\Assembly} &
   \makecell{Bulb\\Installation} &
   \makecell{Peg Reorientation\\(Occluded)} &
   \makecell{Object Search\\(Occluded)} &
   \makecell{Ball Sorting\\(Dim)} \\
\midrule
Vision only
& \(0.14 \pm 0.01\)
& \(0.49 \pm 0.03\)
& \(0.51 \pm 0.02\)
& \(0.63 \pm 0.02\)
& \(0.80 \pm 0.03\)
& \(0.76 \pm 0.02\)
& \(0.29 \pm 0.02\) 
& \(0.52 \pm 0.04\)
& \(0.57 \pm 0.03\) \\
Vision + TacRGB
& \(0.21 \pm 0.02\)
& \(0.49 \pm 0.02\)
& \(0.51 \pm 0.01\)
& \(0.57 \pm 0.02\)
& \(0.78 \pm 0.03\)
& \(0.72 \pm 0.04\)
& \(\mathbf{0.39 \pm 0.02}\)
& \(\mathbf{0.69 \pm 0.01}\)
& \(\mathbf{0.72 \pm 0.02}\) \\
Vision + TacFF
& \(\mathbf{0.40 \pm 0.02}\)
& \(\mathbf{0.56 \pm 0.01}\)
& \(\mathbf{0.58 \pm 0.02}\)
& \(\mathbf{0.66 \pm 0.01}\)
& \(\mathbf{0.81 \pm 0.03}\)
& \(\mathbf{0.81 \pm 0.02}\)
& \(0.28 \pm 0.02\)
& \(0.52 \pm 0.02\)
& \(0.60 \pm 0.02\) \\
\bottomrule
\end{tabular}%
}
\vspace{0.3em}
\end{table*}

\textbf{Results}. 
Table~\ref{tab:modality_bechmarking} reports the success rates of the policy evaluations for the three aforementioned sensing configurations across the nine simulation benchmark tasks. Corresponding learning curves for all tasks under these configurations are shown in Fig.~\ref{fig:training_curves_modality}.
Each reported success rate represents the average of approximately $1500$ evaluations ($3$ random seeds, $50$ environment initializations, and the last $10$ training epochs), ensuring strong statistical significance. Even small percentage-point differences, therefore, reflect meaningful performance gains.
In the insertion category, peg insertion success increases by $26$ percentage points when TacFF is added, showing that force-based tactile feedback is essential for precise alignment and control.
In the screwing category, bulb installation improves by about $5$ percentage points with TacFF, indicating that tactile force cues enhance control during tightening.
For exploration tasks, TacRGB provides the greatest benefit, improving object search by approximately $17$ percentage points and ball sorting under dim lighting by around $15$ percentage points compared to vision only.
These results address the first research question by demonstrating that tactile sensing provides complementary feedback that enables the robot to perform more reliably both contact-rich manipulation tasks, such as insertion and screwing, and perception-challenging exploration tasks, such as peg reorientation, object search, and ball sorting, in visually-degraded environments.

However, not all tactile information contributes equally across task types. The performance gain of each tactile modality depends strongly on the nature of the task and the characteristics of the required manipulation skill. Tasks involving precise contact alignment, geometric fitting, and force regulation, such as insertion and screwing, benefit the most from TacFF feedback. This representation encodes distributed normal and shear forces across the contact surface, providing physically grounded cues that help the policy maintain stable contact and perform fine alignment. In contrast, tasks that rely more on perceiving surface texture and geometry, such as object search or ball sorting, depend more heavily on TacRGB signals. In these cases, high-resolution tactile images capture detailed surface deformation and shape information. For instance, in ball sorting under dim lighting, the success rate increases from $57\%$ (vision only) to $72\%$ with TacRGB, outperforming TacFF ($60\%$). These modality-specific trends address the second research question, showing that the impact of tactile sensing depends not only on its inclusion but also on the type of tactile information and the manipulation context in which it is applied.

That said, adding more sensory data does not always guarantee improved performance. In certain cases, such as the gear assembly task, the vision-only policy ($63\%$) achieves higher performance than the vision + TacRGB configuration ($57\%$). When the additional sensory input does not provide task-relevant or well-structured information, it can introduce noise and distract the policy during training. Complex tactile image patterns may cause the model to overfit or attend to visually detailed but semantically weak features, ultimately degrading performance. In contrast, TacFF provides a more compact and physically meaningful description of contact interaction that the policy can more easily associate with action outcomes. Consequently, integrating TacFF generally leads to consistent improvements, while TacRGB may sometimes reduce performance when the task does not require detailed surface information.

\begin{figure*}[!ht]
    \centering
    \subfloat[Power plug insertion.\label{fig:power_plug_isaac}]{%
        \includegraphics[width=0.48\linewidth]{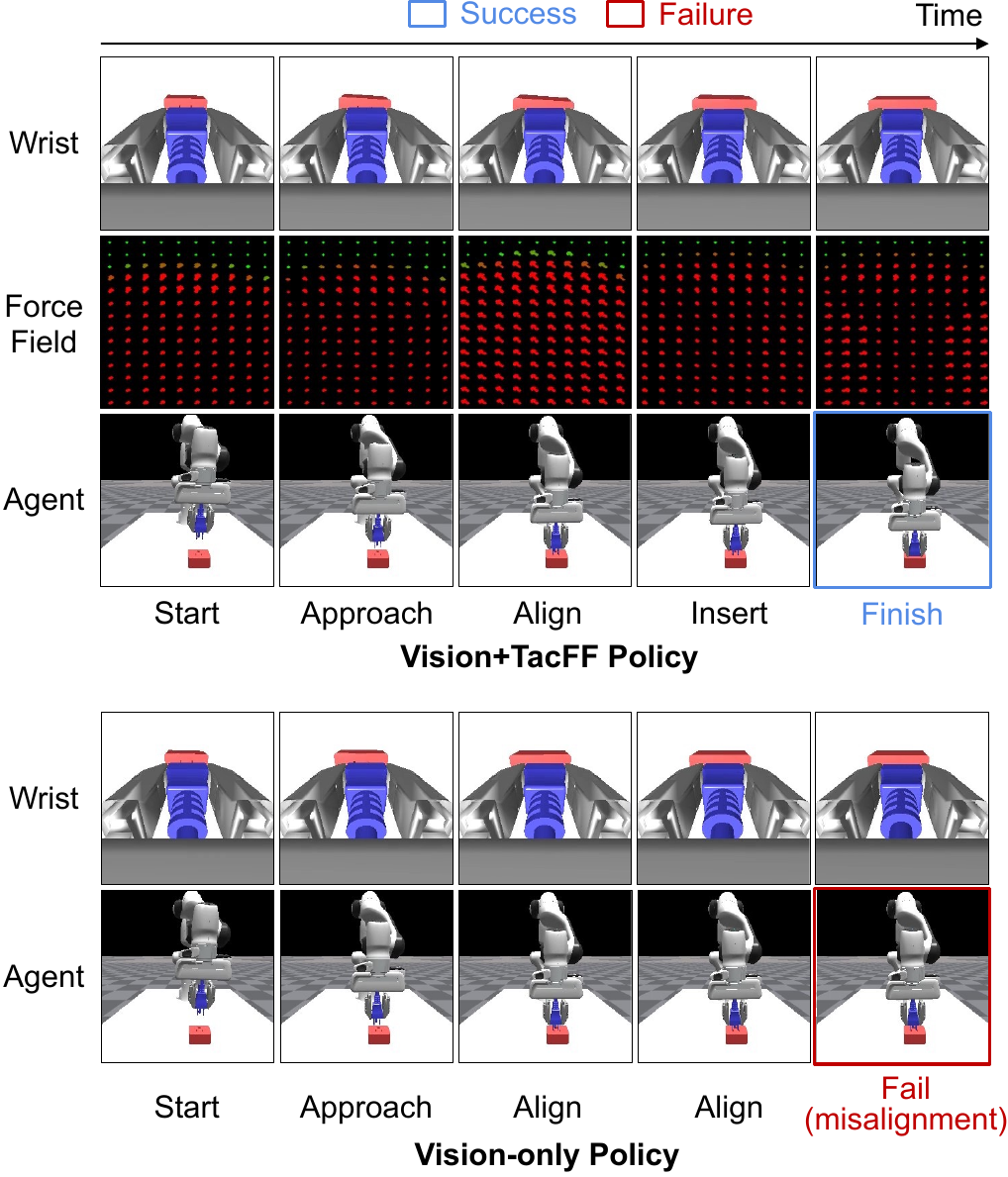}}
    \hfill
    \subfloat[Ball sorting.\label{fig:sorting_isaac}]{%
        \raisebox{2.0pt}{\includegraphics[width=0.48\linewidth]{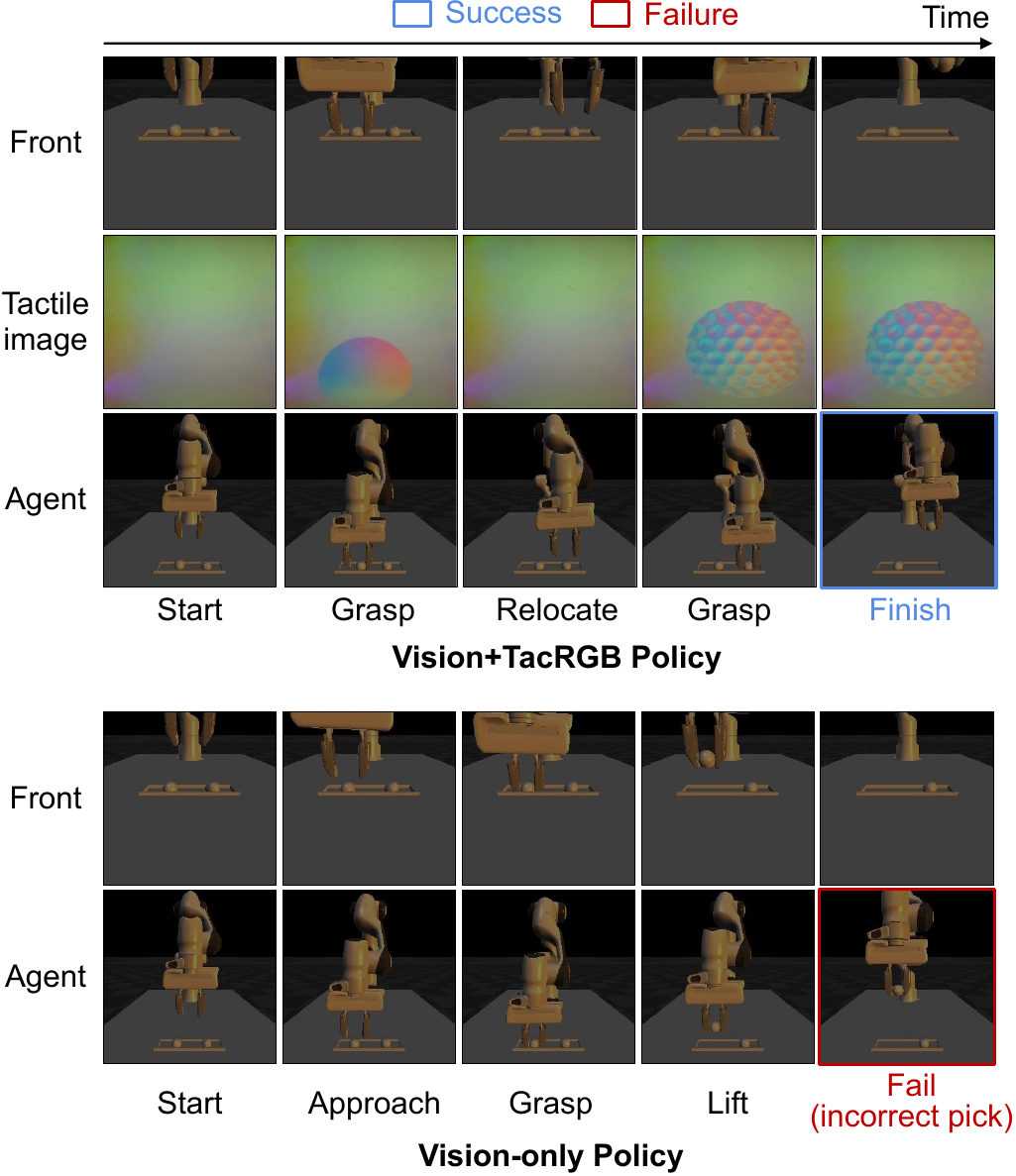}}}
        
    \caption{Representative rollouts of the vision-only and visuotactile policies, including (a) ball sorting and (b) power plug insertion. The typical failure cases of the vision-only method are also shown. The agent-view camera is shown only for visualization, and tactile observations are rescaled for consistent visualization.}
    \label{fig:sim_rollouts}
\end{figure*}

Figure~\ref{fig:sim_rollouts} demonstrates policy rollouts for two representative tasks from the ManiFeel benchmark across vision-only and visuotactile policies: power-plug insertion and ball sorting under dim lighting. These visualizations show the temporal evolution of observations from multiple sensory channels, including camera views and tactile inputs, across key action stages. 
In the power-plug insertion task, the vision-only policy frequently fails due to misalignment during insertion, as the socket becomes partially occluded and visual feedback alone cannot provide precise contact cues. In contrast, the visuotactile policy leverages force-field feedback (TacFF) to detect and correct subtle contact deviations, enabling accurate alignment and successful insertion.
For the ball-sorting task, the vision-only policy often misidentifies balls with similar visual appearances, leading to incorrect picks. The visuotactile policy, equipped with high-fidelity tactile images (TacRGB), distinguishes surface textures, allowing reliable object classification and successful task completion even under dim lighting. Additional examples and full rollouts are provided in the supplementary video.

Together, these results reveal how tactile sensing enhances robotic policy learning in contact-rich or visually ambiguous conditions while demonstrating the importance of selecting and integrating modalities appropriately. They highlight a key challenge for future visuotactile policy learning: effective multimodal integration must emphasize structured, task-relevant tactile representations to fully realize the benefits of touch in robotic manipulation.

\subsection{Policy Benchmarking} \label{sec:policy_benchmark}
\textbf{Setup}. In this section, we benchmark three policy architectures: Diffusion Policy (DP)~\cite{chi2023diffusionpolicy}, Equivariant Diffusion Policy (EquiDiff)~\cite{wang2024equivariant}, and Flow Matching (FM)~\cite{zhang2024flowpolicy}, across four representative manipulation tasks selected from the three task categories: gear assembly, bulb installation, object search, and ball sorting. Each policy is evaluated under the three aforementioned sensing configurations. For tactile input, we use a ResNet18 encoder trained from scratch, while for visual input, DP and FM employ a standard ResNet18 encoder and EquiDiff uses an equivariant ResNet18~\cite{wang2024equivariant}. Additional details on policy architectures and training parameters are provided in Appendices~\ref{app:pipeline_details}--\ref{app:experiment_details}.

\begin{table*}[t]
\centering
\caption{Success rates (mean $\pm$ std) averaged over 3 seeds across tasks and modalities. Results are shown for three policy architectures (DP, EquiDP, FM). Bold indicates the best performance within each task and architecture.}
\label{tab:TactilePolicyBenchmark}
\vspace{1mm}

\resizebox{\textwidth}{!}{
\begin{tabular}{l|ccc|ccc|ccc}
\toprule
 & \multicolumn{3}{c|}{\textbf{DP}} & \multicolumn{3}{c|}{\textbf{EquiDP}} & \multicolumn{3}{c}{\textbf{FM}} \\
\cmidrule(lr){2-4} \cmidrule(lr){5-7} \cmidrule(lr){8-10}
\textbf{Task} &
\textbf{\makecell{Vision\\only}} &
\textbf{\makecell{Vision +\\TacRGB}} &
\textbf{\makecell{Vision +\\TacFF}} &
\textbf{\makecell{Vision\\only}} &
\textbf{\makecell{Vision +\\TacRGB}} &
\textbf{\makecell{Vision +\\TacFF}} &
\textbf{\makecell{Vision\\only}} &
\textbf{\makecell{Vision +\\TacRGB}} &
\textbf{\makecell{Vision +\\TacFF}} \\
\midrule
Gear Assembly &
0.63 $\pm$ 0.02 & 0.57 $\pm$ 0.02 & \textbf{0.66 $\pm$ 0.01} &
0.70 $\pm$ 0.02 & 0.66 $\pm$ 0.02 & \textbf{0.74 $\pm$ 0.03} &
0.38 $\pm$ 0.01 & 0.37 $\pm$ 0.01 & \textbf{0.39 $\pm$ 0.02} \\
Bulb Installation &
0.76 $\pm$ 0.02 & 0.72 $\pm$ 0.04 & \textbf{0.81 $\pm$ 0.02} &
0.94 $\pm$ 0.01 & 0.94 $\pm$ 0.01 & \textbf{0.95 $\pm$ 0.02} &
0.60 $\pm$ 0.03 & 0.63 $\pm$ 0.03 & \textbf{0.66 $\pm$ 0.02} \\
Object Search &
0.52 $\pm$ 0.04 & \textbf{0.69 $\pm$ 0.01} & 0.52 $\pm$ 0.02 &
0.44 $\pm$ 0.08 & \textbf{0.67 $\pm$ 0.04} & 0.46 $\pm$ 0.07 &
0.16 $\pm$ 0.03 & \textbf{0.44 $\pm$ 0.03} & 0.15 $\pm$ 0.03 \\
Ball Sorting &
0.43 $\pm$ 0.03 & \textbf{0.73 $\pm$ 0.03} & 0.50 $\pm$ 0.03 &
0.21 $\pm$ 0.03 & \textbf{0.46 $\pm$ 0.02} & 0.31 $\pm$ 0.04 &
0.41 $\pm$ 0.03 & \textbf{0.69 $\pm$ 0.01} & 0.43 $\pm$ 0.03 \\
\bottomrule
\end{tabular}%
}
\end{table*}

\begin{figure*}[t]
    \centering
    \subfloat[Gear Assembly\label{fig:modality_gear}]{%
        \includegraphics[width=0.49\linewidth]{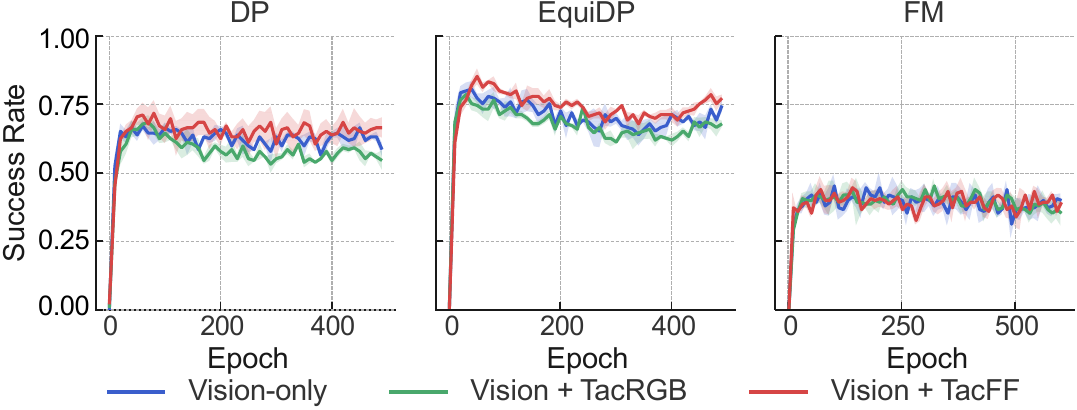}}
    \hfill
    \subfloat[Bulb Installation\label{fig:modality_bulb}]{%
        \includegraphics[width=0.49\linewidth]{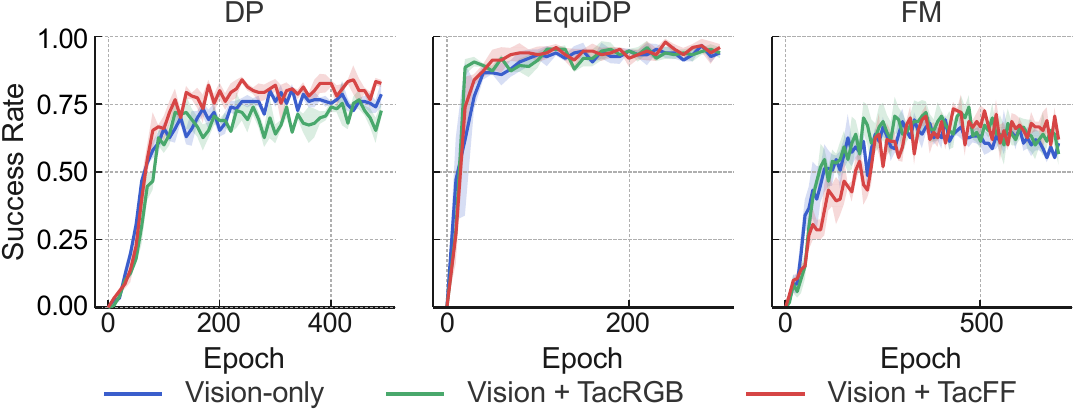}}
    
    \subfloat[Object Search\label{fig:modality_search}]{%
        \includegraphics[width=0.49\linewidth]{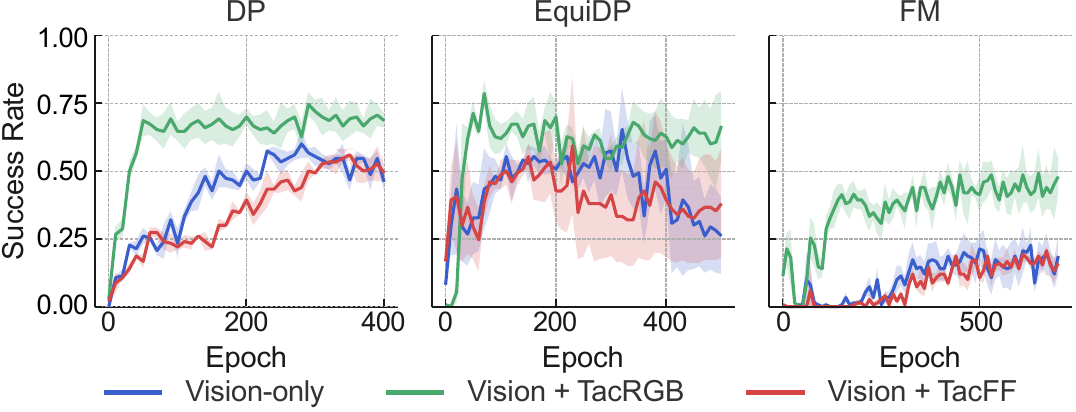}}
    \hfill
    \subfloat[Ball Sorting\label{fig:modality_sort}]{%
        \includegraphics[width=0.49\linewidth]{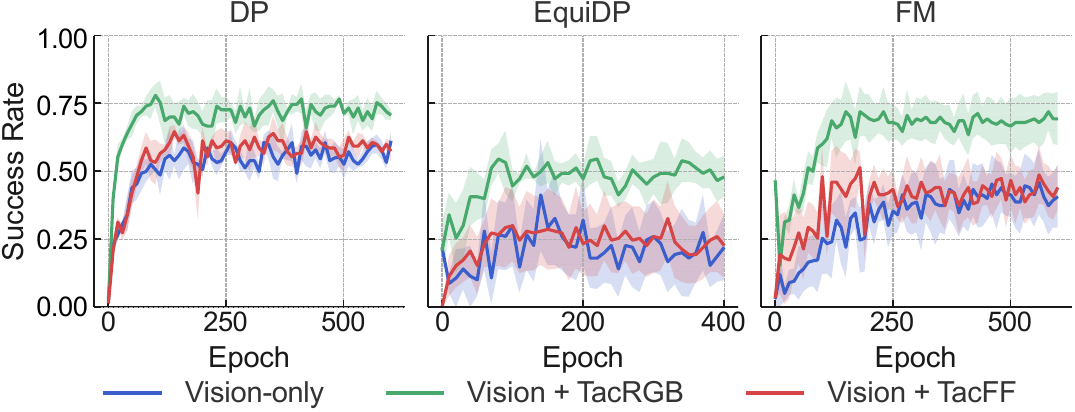}}
        
    \caption{Learning curves comparing the three different sensing configurations (i.e., Vision-only, Vision + TacRGB, and Vision + TacFF) across different policy architectures, for four representative tasks: (a) Gear Assembly, (b) Bulb Installation, (c) Object Search, and (d) Ball Sorting}
    \label{fig:training_curves_policy}
\end{figure*}

\textbf{Results}. As shown in Table~\ref{tab:TactilePolicyBenchmark}, all three policy architectures exhibit consistent trends across tasks when tactile sensing is incorporated. The corresponding learning curves for these policy configurations are provided in Fig.~\ref{fig:training_curves_policy}. Adding TacFF generally improves performance in contact-rich insertion and screwing tasks, while TacRGB provides greater benefit in exploration tasks where visual input is limited. For instance, in gear assembly, TacFF increases success by about $3$ percentage points for DP and $4$ percentage points for EquiDP compared to their vision-only baselines, reflecting the importance of force-based tactile feedback for precise alignment and contact regulation. In contrast, TacRGB yields the highest gains in object search and ball sorting, improving EquiDP vision-only performance by $23$ and $25$ percentage points, and FM vision-only performance by $28$ percentage points in both tasks, respectively. These improvements confirm the advantage of high-resolution tactile images in providing detailed surface information that compensates for degraded visual cues.

Across architectures, FM performs worse than DP and EquiDP in most benchmarking tasks. Between the diffusion-based methods, EquiDP consistently achieves higher success rates in insertion and screwing tasks, such as bulb installation and gear assembly. This advantage likely arises because these tasks rely heavily on visual feedback for coarse localization and alignment before tactile sensing becomes critical for fine contact adjustment. The equivariant encoder in EquiDP effectively captures spatial symmetries from wrist-mounted camera observations, allowing it to leverage visual geometry more efficiently. In such tasks, improvements in vision-only performance often translate proportionally to gains in visuotactile policies.
In contrast, exploration tasks, including object search and ball sorting, depend more on tactile sensing for object recognition and classification under occluded or low-light conditions. In these cases, EquiDP offers limited benefits, as its strength in visual geometric reasoning is less applicable to the tactile-driven fusion required for these scenarios. As a result, EquiDP sometimes performs slightly worse than the standard DP, which demonstrates stronger robustness when visual cues are unreliable and achieves more effective fusion of high-fidelity tactile sensing for visuotactile policies.

Additionally, as discussed in the modality benchmarking results in Section~\ref{sec:modality-benchmark}, adding TacRGB may negatively affect performance when its high-dimensional deformation images provide limited task-relevant information, particularly in contact-rich manipulation tasks. A similar effect appears in bulb installation, where the drop is most pronounced for DP. This suggests that DP is more sensitive to the variability introduced by TacRGB, whereas EquiDP and FM exhibit greater robustness due to their equivariant spatial encoding and smoother optimization dynamics. However, this robustness does not override a fundamental modality-task mismatch. When the tactile modality does not supply the type of information the task relies on, architectural differences may reduce sensitivity but cannot guarantee improved performance or fully eliminate degradation, for example, as shown in the gear assembly task where all architectures still suffer degradation. This highlights the importance of prioritizing the selection of tactile modalities whose sensing characteristics align closely with the demands of the manipulation skill before considering policy architecture design.

Overall, these results confirm that tactile sensing enhances performance across architectures and tasks, but the extent of improvement depends on both the tactile modality and the policy design.


\subsection{Tactile Representation Benchmarking} \label{sec:tac-rep-benchmark}

\textbf{Setup.} We present simulation benchmarking results for vision–TacRGB policies equipped with different tactile representations. These tactile representations encode high-resolution tactile images (TacRGB) into feature-rich embeddings. The evaluation is conducted across three representative manipulation tasks: gear assembly, object search, and ball sorting. The ResNet18 encoder is trained from scratch, while UniT~\cite{xu2025unit}, T3~\cite{zhao2024transferable}, and AnyTouch~\cite{feng2025anytouch} are pre-trained tactile encoders that are fine-tuned during policy training. Further details on the encoder architectures and training parameters are provided in Appendices~\ref{app:pipeline_details}--\ref{app:experiment_details}.

\begin{table}[t]
\centering
\caption{Success rates (mean $\pm$ std) for different tactile representation encoders across representative manipulation tasks in simulation. Results are averaged over three seeds. Best performance in each task is highlighted in bold.}
\label{tab:TactileRepr}

\resizebox{\columnwidth}{!}{%
\begin{tabular}{l|cccc}
\toprule
\textbf{Task} & \textbf{ResNet18} & \textbf{UniT} & \textbf{T3} & \textbf{AnyTouch} \\
\midrule
Gear Assembly           & $0.57 \pm 0.02$ & $\mathbf{0.61 \pm 0.01}$ & $0.57 \pm 0.02$ & $0.28 \pm 0.02$ \\
Object Search     & $0.69 \pm 0.01$ & $0.58 \pm 0.03$ & $\mathbf{0.71 \pm 0.03}$ & $0.59 \pm 0.02$ \\
Ball Sorting         & $0.73 \pm 0.03$ & $\mathbf{0.80 \pm 0.05}$ & $0.68 \pm 0.01$ & $0.74 \pm 0.02$ \\
\bottomrule
\end{tabular}%
}
\vspace{0.3em}
\end{table}

\begin{figure}[t]
    \centering
      \includegraphics[width=\linewidth]{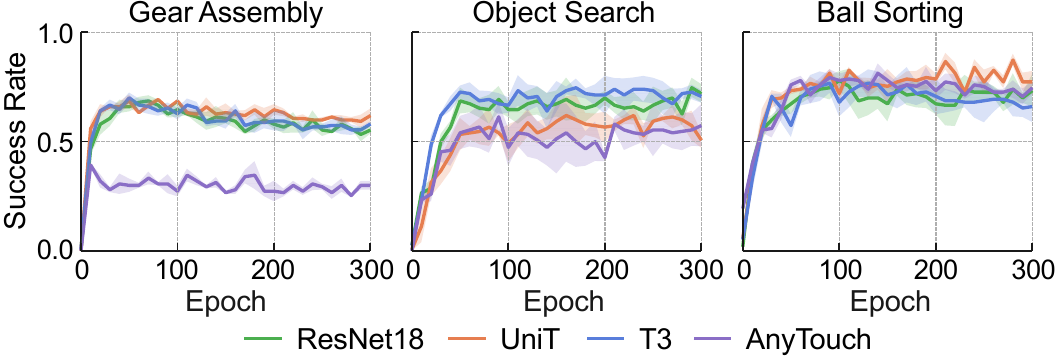}
    \caption{Learning curves across vision--TacRGB policies using different tactile representations, for three representative tasks, namely gear assembly, object search, and ball sorting.}
    \label{fig:training_curves_tactile_repr}
\end{figure}

\textbf{Results.}
Table~\ref{tab:TactileRepr} reports the success rates of vision-TacRGB policies with different tactile representation encoders across the three representative tasks. The corresponding learning curves are shown in Fig.~\ref{fig:training_curves_tactile_repr}. The results indicate that pre-trained tactile encoders such as UniT and T3, which are specifically designed for high-resolution tactile image processing, consistently outperform the baseline ResNet18 trained from scratch. This improvement is especially pronounced in tasks requiring dynamic contact reasoning or texture-dependent recognition, such as object search and ball sorting, where UniT and T3 achieve the highest success rates.
Among the evaluated representations, UniT shows the most balanced performance across all tasks, including the force-sensitive gear assembly. AnyTouch demonstrates strong performance in texture-based reasoning tasks but struggles in tasks that demand fine force-sensitive control, suggesting that current pre-trained tactile encoders may still be specialized toward specific sensing contexts.

Overall, the results emphasize the importance of leveraging pre-trained tactile representations to enhance visuotactile policy learning. However, the performance variation across tasks also reveals that a unified tactile encoder capable of generalizing across diverse manipulation scenarios remains an open challenge. While pretrained visual encoders have demonstrated strong transferability in policy learning~\cite{chi2024universal, pari2021surprising, barreiros2025careful}, this trend does not consistently extend to tactile representations. Although several tactile representation models have been specifically designed for policy learning, their performance remains inconsistent across tasks and modalities. This highlights the ongoing need for more robust and generalizable tactile representation models that can effectively support policy learning in diverse contact conditions. Developing such representations that can seamlessly adapt to both spatial contact interactions and texture-dependent tactile cues will be crucial for advancing robust visuotactile robot learning. We envision ManiFeel as a reproducible testbed to facilitate this progress, similar to how standardized benchmarks in vision have accelerated the development of transferable visual representations.

\subsection{Real-world Evaluation}\label{sec:real_world_eval}

\textbf{Setup.} To assess the validity of the proposed ManiFeel simulation benchmark, we examine the consistency between simulated and real-world performance across three representative tasks: gear assembly, bulb installation, and ball sorting under both normal and dim lighting conditions.
All policies are trained from scratch using real-world datasets collected specifically for the real-world settings. We benchmark the three sensing configurations introduced earlier (vision only, vision + TacRGB, and vision + TacFF), using ResNet18 encoders trained from scratch and Diffusion Policy as the policy backbone.
Experiments are conducted using a Franka Emika Panda manipulator\footnote{\url{https://franka.de/products/franka-research-3}} equipped with a parallel gripper integrated with a GelSight R1.5 tactile sensor~\cite{wang2021gelsight}. In the real setup, the tactile RGB signal ($\mathbf{I}^{\text{tac}}$) is directly captured as raw high-resolution images from the GelSight R1.5 sensor, providing detailed surface deformation patterns within the contact region. To obtain the tactile force-field map ($\mathbf{S}^{\text{ff}}$), we estimate local contact forces by combining marker-based optical flow and depth reconstruction following~\cite{wang2021gelsight}. Specifically, two-dimensional marker displacements encode tangential shear components, while the reconstructed surface depth provides the normal indentation magnitude, together forming a spatial map that encodes distributed normal and shear forces across the contact surface. Both tactile modalities are normalized to the range $[-1, 1]$ before being passed to their respective encoders, ensuring consistency in dynamic range between real and simulated TacRGB and TacFF signals. 
Further details on the task setup and training setup are provided in Appendices~\ref{app:task_details} and~\ref{app:experiment_details}.

\begin{figure*}[t]
    \centering

    \subfloat[Gear assembly.\label{fig:real_gear_process}]{%
        \includegraphics[width=0.49\linewidth]{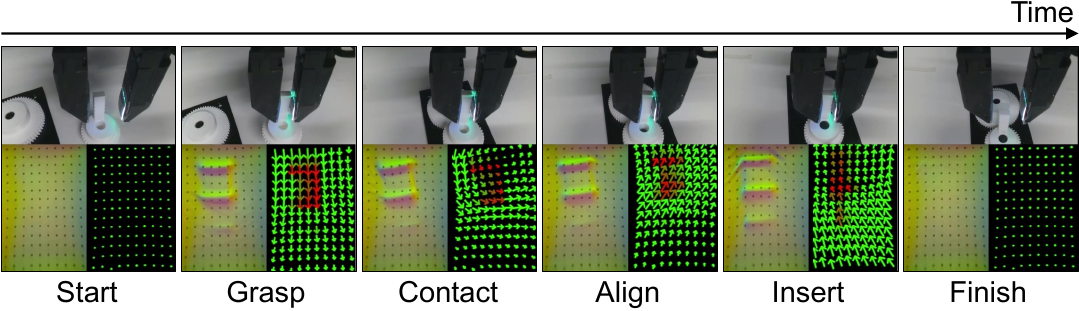}}
    \hfill
    \subfloat[Bulb installation.\label{fig:real_bulb_process}]{%
        \raisebox{1.4pt}{\includegraphics[width=0.49\linewidth]{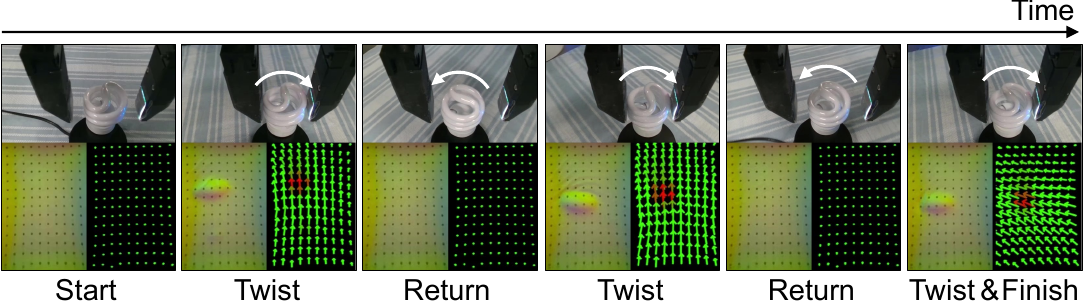}}}
    
    \vspace{-4mm}
    \subfloat[Ball sorting under normal lighting.\label{fig:real_sorting_process_normal}]{%
        \includegraphics[width=0.49\linewidth]{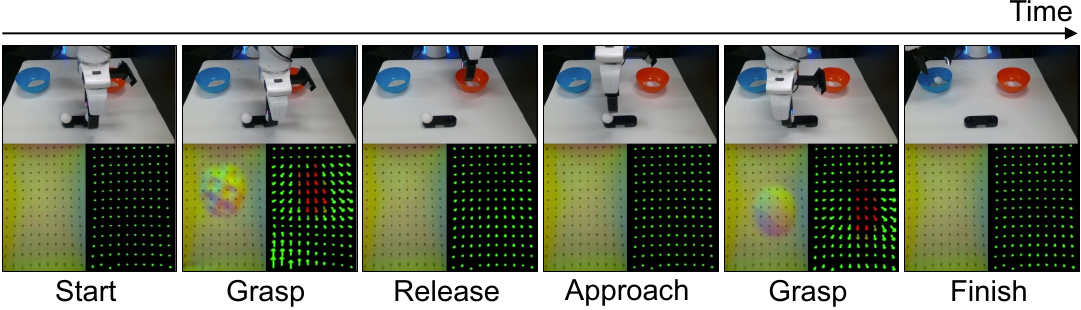}}
        \hfill
    \subfloat[Ball sorting under dim lighting.\label{fig:real_sorting_process_dim}]{%
        \raisebox{0.2pt}{\includegraphics[width=0.49\linewidth]{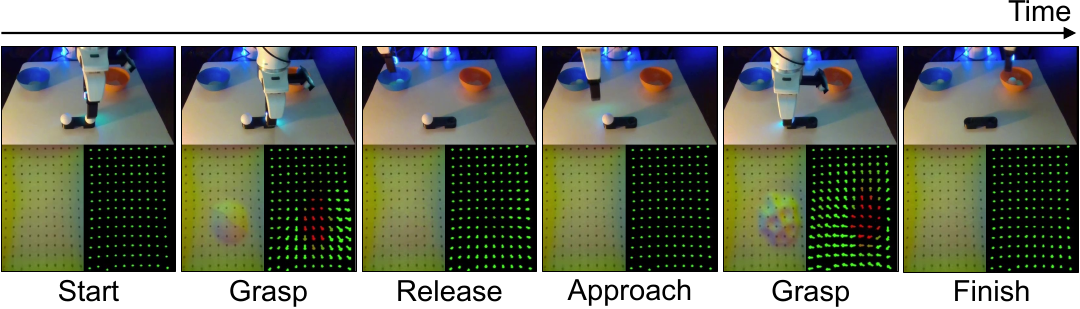}}}
    
    \caption{Demonstrations of teleoperated trajectories in real-world tasks: (a) gear assembly, (b) bulb installation, (c) ball sorting under normal lighting, and (d) ball sorting under dim lighting. Visual observations are captured from a wrist-mounted camera for (a) and (b) and from a front-facing camera for (c) and (d). All tactile observations are recorded using the right-finger GelSight R1.5 sensor.}

    \label{fig:real-tac-sorting-rollout}
\end{figure*}

\textbf{Representative Teleoperated Rollouts.}
Fig.~\ref{fig:real-tac-sorting-rollout} presents several demonstrations of teleoperated trajectories along with corresponding visual and tactile observations for the three representative tasks.
In gear assembly, the wrist-mounted camera provides limited visibility of the contact interface, making it difficult to determine whether the gear teeth are properly aligned. Tactile feedback, particularly from the tactile force field, helps detect misalignment and guide corrective adjustments for successful insertion.
In bulb installation, visual input alone cannot reliably indicate whether the bulb has been sufficiently tightened, which may result in over- or under-tightening. Tactile sensing provides complementary feedback that allows the policy to identify the correct tightening point.
In tactile sorting, distinguishing between visually similar spherical objects is challenging using vision alone. High-fidelity tactile images capture the differences in surface texture, enabling accurate identification and manipulation of each object. Autonomous policy rollouts, including both successful executions and representative failure cases for each task, are provided in the accompanying supplementary video.

\begin{figure*}[t]
    \centering
    \includegraphics[width=0.90\linewidth]{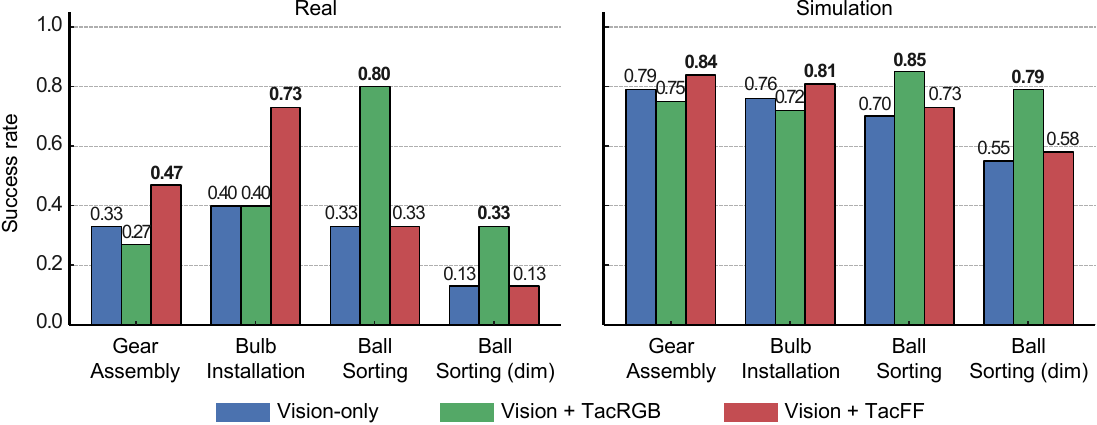}
    \caption{Success rates for real and simulated environments across different sensing configurations and scenarios.}
    \label{fig:real_vs_sim_results}
\end{figure*}



\textbf{Results.}
Figure~\ref{fig:real_vs_sim_results} shows the comparison between simulated and real environments across three sensing configurations. The overall trends observed in simulation are consistently reproduced in the real world, demonstrating that ManiFeel’s task suite captures representative and transferable manipulation behaviors. In both domains, incorporating tactile sensing substantially improves policy performance, particularly in contact-rich and visually constrained scenarios. For instance, in gear assembly, TacFF improves success by $14$ percentage points over the vision-only baseline (from $33\%$ to $47\%$) in real experiments, reflecting similar improvements in simulation. Likewise, TacFF achieves the highest success in bulb installation ($73\%$ in real, $81\%$ in sim), confirming the importance of force-based tactile feedback for accurate tightening and contact regulation.

For the ball-sorting task, both normal and dim-light conditions exhibit consistent trends with simulation, where incorporating TacRGB improves policy performance. In particular, TacRGB enables more reliable object perception and task execution when visual cues are degraded or insufficient, as in the ball-sorting scenario where the balls appear similar in shape and size. This improvement in manipulation performance is significant, especially under normal lighting, where the real policy’s success increases from $33\%$ to $80\%$, matching the improvement trend observed in simulation ($70\%$ to $85\%$). Under dim lighting, performance drops mainly because the vision module struggles to localize and grasp the ball before tactile feedback becomes available. Once contact is made, however, TacRGB still enables stable ball classification and accurate transport, yielding a relative gain consistent with simulation and demonstrating its effectiveness under visually limited conditions.

Overall, these results confirm that ManiFeel produces consistent and interpretable performance trends between simulation and reality, validating that our simulation benchmark effectively reflects the nature of real-world visuotactile manipulation. While reproducing identical quantitative results across domains is inherently impractical and beyond the scope of this work, the consistency of relative trends demonstrates that ManiFeel provides a reliable and meaningful foundation for benchmarking and future visuotactile policy development.
\section{Discussion and Limitations} \label{sec:discussion}
\subsection{Discussion and Insights}
ManiFeel provides a unified and extensible framework for systematically evaluating visuotactile policy learning and investigating how tactile sensing enhances robotic manipulation across contact-rich and visually constrained environments. Our results yield several key insights that directly address the research questions posed in this study.

First, across both simulation and real-world experiments, tactile sensing consistently improves policy performance in contact-rich manipulation and in scenarios where visual input is limited or ambiguous. These results confirm that tactile sensing provides complementary feedback that is essential for robust control in contact-dominated or visually uncertain environments.

Second, the effectiveness of each tactile modality depends strongly on the task’s characteristics. The tactile force field (TacFF), which captures distributed normal and shear forces, excels in tasks involving geometric fitting and force-sensitive interactions, such as insertion and screwing. In contrast, high-fidelity tactile images (TacRGB) are more effective for tasks requiring recognition of object pose and surface texture, such as peg reorientation, object search, and ball sorting, particularly in settings where visual cues are limited. These modality-specific patterns reveal that the contribution of tactile sensing is highly task dependent; its impact is determined by how well the tactile representation aligns with the dominant physical interaction cues of the manipulation context.

Third, results from real-world experiments closely mirror simulation trends, validating the fidelity of ManiFeel in capturing visuotactile interaction dynamics. Across domains, TacFF consistently enhances performance in force-sensitive assembly tasks, while TacRGB improves object perception in visually degraded conditions. However, quantitative discrepancies remain, particularly under severe lighting degradation, where real camera noise and reduced contrast hinder visual localization. Differences in tactile signal quality, elastomer deformation, and frictional dynamics between simulated and physical sensors also contribute to these performance gaps. These findings highlight ManiFeel’s value as a high-fidelity yet statistically meaningful simulation benchmark, while emphasizing the need for continued progress in bridging sim-to-real disparities in visuotactile sensing.

Lastly, our benchmarking results suggest several practical insights for future visuotactile policy design.
\textbf{(i) Task-specific modality selection is crucial:} TacFF should be prioritized for contact-rich manipulation and force-sensitive tasks like insertion and screwing, while TacRGB is more effective in tasks that require sensing of object geometry and surface textures, particularly under visually uncertain conditions. If modality selection is not carefully considered, adding additional sensory inputs can degrade learning when the added information is misaligned or uninformative to the task characteristics, as observed in gear assembly with TacRGB.
\textbf{(ii) Policy architecture design matters across scenarios:} Even with similar fusion strategies and observation encoders, the design of the action head plays an important role. For example, DP outperforms FM under identical encoder and fusion settings, indicating the advantage of the denoising-based action generation mechanism. Likewise, when the same action head is used, enhancing the observation encoder, such as adopting an equivariant visual encoder in EquiDP, can substantially improve visuotactile policy performance. This improvement is especially evident in contact-rich tasks like insertion and screwing, where global visual information is critical. On the other hand, the results also show that equivariant visual encoders may underperform in tactile-centric environments with uncertain or unreliable visual cues, such as occluded object search or dimly lit sorting, suggesting careful consideration of visual encoder design in such environments. 
\textbf{(iii) Tactile representation quality matters:} Pretrained tactile encoders such as UniT and T3 achieve higher performance than task-specific CNNs. This demonstrates the benefit of large-scale pretraining and the importance of robust tactile representations for reliable visuotactile policy learning. However, more effective tactile representation models are still needed, as the performance of existing methods remains inconsistent across different tasks.

\subsection{Limitations and Future Work}
While ManiFeel provides a unified and extensible framework for benchmarking visuotactile policy learning, several limitations remain that motivate future research directions. 

First, our results reveal that existing tactile representations do not generalize consistently across tasks. Pretrained encoders such as UniT and T3 improve performance on certain tasks but fail to transfer robustly to others, indicating the need for more universal tactile representations that can adapt to diverse contact dynamics and task objectives. Although ManiFeel currently benchmarks a representative set of tactile encoders, we acknowledge the existence of other tactile representation methods, such as Sparsh \cite{higuera2024sparsh}, which we plan to explore in future work. While a comprehensive evaluation of all available representations is beyond the current scope due to limited computational and experimental resources, ManiFeel is designed with a modular and open structure that enables future studies to easily extend the benchmark with additional tactile encoders, representations, and policy architectures.

Second, ManiFeel focuses on pairwise combinations of vision with a single tactile modality at a time (either TacRGB or TacFF) and employs a concatenation-based visuotactile fusion baseline to isolate the contribution of individual modalities. This design choice enables clear attribution of each sensory
input and facilitates analysis of how different tactile representations influence policy performance. Although combining vision with both TacRGB and TacFF could potentially improve overall task performance across scenarios, such a configuration may not necessarily outperform the best pairwise combination and would make it more difficult to clearly interpret the contribution of each modality. Moreover, ManiFeel does not yet incorporate more advanced multimodal fusion mechanisms. The observed variation in performance across tasks highlights the importance of developing adaptive, task-aware fusion architectures that can dynamically weight modalities according to their relevance. ManiFeel provides a standardized foundation and compatible training pipelines that can facilitate future investigations into cross-modal representation learning and hierarchical fusion strategies.

Third, for policy benchmarking, tactile representation evaluation, and real-world experiments, we focus on representative tasks from each category to maintain experimental feasibility while enabling systematic comparison and in-depth analysis. Although this limits full task coverage, the selected tasks capture the key contact dynamics within each category and provide insights that generalize reasonably well to similar manipulation tasks.

Finally, although real-world experiments largely confirm the performance trends observed in simulation, some discrepancies remain in absolute success rates. These differences arise from the limited fidelity of simulated tactile signals, camera sensitivity, and unmodeled contact dynamics. The real-world setups also differ slightly from their simulated counterparts, particularly for the ball-sorting task, since the goal of this work was to validate overall consistency rather than perform direct policy transfer. Future efforts will aim to reduce this gap through improved tactile sensor modeling, domain adaptation, and randomization.

\section{Conclusion}
This work presents \textbf{ManiFeel}, a simulation benchmark designed to advance supervised visuotactile policy learning. Through extensive experiments across diverse manipulation tasks, we systematically evaluate the influence of visuotactile sensing modalities, tactile representations, and policy architectures on performance, providing key insights into effective visuotactile policy design. The results highlight the critical role of tactile sensing in both contact-rich manipulation and scenarios where visual cues are limited, as well as the ongoing challenges in developing generalizable tactile encoders and advanced visuotactile policy architectures. Finally, ManiFeel establishes a reproducible and extensible foundation for visuotactile policy benchmarking. Its modular design encourages the community to explore new tactile representations, policy architectures, and multimodal fusion strategies, advancing toward more robust and generalizable visuotactile robotic manipulation.

\appendices
\section{Additional Details on Task Environments} \label{app:task_details}

This section provides additional environment configuration details for each task used in our real-world and simulation experiments. All task setups are built upon the IsaacGym simulation\footnote{https://github.com/isaac-sim/IsaacGymEnvs.git}. The code and configuration files for all simulation task setups will be open-sourced to facilitate reproducibility.

\textbf{Insertion Tasks (Sim).}
For the peg, USB, power plug, and gear assembly tasks, both the socket location and the in-hand orientation of the object (peg, USB, plug, or insertion gear) are randomized at the start of each rollout. The socket position or gear base is sampled within a ($-5$\,cm, $+5$\,cm) range along the $x$ and $y$ axes, and the object’s in-hand orientation is randomized within $10^\circ$ of its nominal pose. The gripper remains closed throughout to maintain a stable grasp during precise alignment and insertion (see Fig.~\ref{fig:sim_demo_trajectory_insertion}).

\textbf{Screwing Tasks (Sim).}
This category includes nut–bolt assembly and bulb installation. For both tasks, the bolt or socket position is randomized within ($-5$\,cm, $+5$\,cm) along the $x$ and $y$ axes from its nominal location at the start of each rollout, while the in-hand rotation of the object (nut or bulb) is fixed initially. The robot end-effector's roll and pitch angles are also fixed during execution. In this setup, the robot hand first approaches the bolt or socket and then performs the screwing motion. At this stage, the gripper’s open–close action and applied grasping force are regulated to maintain stable contact and ensure reliable threading.

\textbf{Peg Reorientation (Sim).}
The robot must perform self-alignment by leveraging external contact with the socket rim to rotate the peg in hand before insertion into a socket placed inside an enclosed container, emulating how humans manipulate or orient objects in confined spaces without visual access. The socket position remains fixed within a box, while the peg’s initial in-hand orientation is randomized up to $30^\circ$.  The gripper remains closed throughout to ensure a stable grasp during in-hand reorientation and insertion (see Fig.~\ref{fig:peg_reorientation_process}).

\textbf{Object Search (Sim).}
This task requires the robot to pick up a cube-shaped battery from inside a closed container, emulating how humans retrieve objects from a bag without visual access. At the beginning of each demonstration or evaluation, the object’s position is randomly selected between two predefined locations along the $x$-axis within the container. Because the interior is fully occluded, the robot relies solely on tactile feedback to explore, localize, and grasp the object (see Fig.~\ref{fig:object_search_process}).

\textbf{Ball Sorting (Sim).}
This task requires the robot to pick up the golf ball from a pair of golf and table-tennis balls placed under dim lighting conditions to simulate degraded visual perception. A golf ball and a table tennis ball are placed on a ball stand whose position is randomized within ($-5$\,cm, $+5$\,cm) along the $x$ and $y$ axes from its nominal position. The ball positions (golf vs. table tennis) are swapped across episodes to increase task diversity (see Fig.~\ref{fig:ball_sorting_process}).

\textbf{Gear Assembly (Real).}
In the real-world setting, the robot first grasps the insertion gear and then aligns and mounts it onto the central shaft of a base featuring two side gears. During data collection and evaluation rollouts, the gear base is slightly rotated in-plane from its nominal orientation to introduce variability (see Fig.~\ref{fig:real_gear_process}).

\textbf{Bulb Installation (Real).}
In this setup, the bulb is initially placed in its socket, and the robot moves to grasp and screw it into place. The initial screw depth of the bulb within the socket is randomized across rollouts to introduce variability, resulting in uncertainty in the number of turns required for proper tightening (see Fig.~\ref{fig:real_bulb_process}).

\textbf{Ball Sorting (Real).}
The robot must classify and sort two visually similar spherical objects, a golf ball and a table tennis ball, into orange and blue bowls, respectively. The task is conducted under normal and dim lighting conditions. The positions of the two balls are swapped between rollouts to introduce variation (see Figs.~\ref{fig:real_sorting_process_dim}--\ref{fig:real_sorting_process_normal}).

\section{Dataset Details} \label{app:dataset_details}

We collect expert demonstrations using a 3Dconnexion SpaceMouse for all tasks in the ManiFeel benchmark, including $9$ simulation and $4$ real-world tasks. Table~\ref{tab:dataset_summary} summarizes the amount of demonstrations, camera views, tactile sensors, and action dimensions (ActD) recorded per task. Tasks with an action dimension of $7$ include gripper open/close actions, while those with an ActD of $6$ use a fixed gripper state, which is excluded from the action space. In the nut--bolt assembly, bulb installation, and object search tasks, although our dataset records ActD of $7$ for pipeline consistency, the end-effector rotations are constrained during execution, where roll–pitch angles are fixed for the nut--bolt assembly and bulb installation, and roll–pitch–yaw angles are fixed for the object-search task. For real-world experiments, the roll and pitch angles are fixed throughout execution, resulting in an ActD of $5$.

\begin{table}[ht]
  \centering
  \caption{\textbf{Dataset Summary}. Cam: number of camera viewpoints recorded in the dataset; Tac: number of tactile sensors recorded; PropD: proprioceptive feedback dimension; ActD: action dimension; Demo: number of human demonstrations collected.}
  \label{tab:dataset_summary}
  \vspace{1mm}
  \begin{tabular}{l|*{6}{c}}
    \toprule
    \textbf{Task} & \textbf{Cam} & \textbf{Tac} & \textbf{PropD} & \textbf{ActD} & \textbf{Demo}  \\
    \midrule
    \multicolumn{6}{c}{Simulation Dataset} \\
    \midrule
    Peg Insertion           & 3 & 2 & 7 & 6 & 50 \\
    USB Insertion           & 3 & 2 & 7 & 6 & 50 \\
    Power Plug Insertion    & 3 & 2 & 7 & 6 & 50 \\
    Gear Assembly           & 3 & 2 & 7 & 6 & 100 \\
    Nut--bolt Assembly      & 3 & 2 & 7 & 7 & 30 \\
    Bulb Installation      & 3 & 2 & 7 & 7 & 20 \\
    Peg Reorientation         & 3 & 2 & 7 & 6 & 50 \\
    Object Search         & 3 & 2 & 7 & 7 & 50 \\
    Ball Sorting (Normal)     & 3 & 2 & 7 & 7 & 80 \\
    Ball Sorting (Dim)     & 3 & 2 & 7 & 7 & 80 \\
    
    \midrule
    \multicolumn{6}{c}{Real-world Dataset} \\
    \midrule
    Gear Assembly & 1 & 1 & 7 & 5 & 80 \\
    Bulb Installation & 1 & 1 & 7 & 5 & 50 \\
    Ball Sorting (Normal) & 1 & 1 & 7 & 5 & 80 \\
    Ball Sorting (Dim)    & 1 & 1 & 7 & 5 & 80 \\
    \bottomrule
  \end{tabular}
\end{table}

Each simulation task is recorded with three RGB camera views (front, side, wrist) and two tactile sensors (left and right fingers), whereas the real-world setup includes a front-facing Intel RealSense camera and a single GelSight R1.5 tactile sensor on the right finger. The demonstrations are recorded at $10\,$Hz and include synchronized visual, tactile, and proprioceptive observations. The proprioceptive feedback includes the end-effector’s position and orientation, resulting in a 7-dimensional state vector, where orientation is represented using a quaternion.
While not all camera views and tactile sensors are required as inputs during policy training, our dataset captures diverse sensor observations to support flexible use in future research. The specific observation setup used for our experiments is described in Appendix~\ref{app:experiment_details}. 


\section{Pipeline Details} \label{app:pipeline_details}

\subsection{Visual and Tactile Encoders}
We adopt a ResNet18 CNN encoder (without pretraining) with two key modifications as described in \cite{chi2023diffusionpolicy}: (1) the global average pooling layer is replaced with a spatial softmax pooling layer to retain spatial information; and (2) Batch Normalization is replaced with Group Normalization to ensure stable training. Each camera view is processed using a separate encoder. For visuotactile policies without pretrained tactile representations, we use the same ResNet18 encoder to extract latent embeddings from both tactile images and tactile force fields.

\subsection{Tactile Representations}
\textbf{UniT}  \cite{xu2025unit} is a tactile representation model that can be trained on a small set of simple objects while generalizing to unseen objects. It adopts the training paradigm of VQGAN. During inference, the quantization layer is removed from the encoding process, and only the CNN-based downsampling is used for feature extraction. The resulting UniT representation exhibits a well-structured latent space that emphasizes informative tactile features. 

UniT is not a representation pretrained on broad datasets, but rather a method designed for domain generalization using only a small amount of task-specific data. The checkpoint released by UniT \cite{xu2025unit} was trained on GelSight images with markers, which differ from the tactile observations used in ManiFeel. Therefore, we trained customized UniT representations using data of our domain. For the simulation benchmark, we trained UniT using a set of images of nut and bolt collected through TacSL \cite{akinola2024tacsl}, and used the same representation across all simulation experiments. For real-world experiments, we trained UniT with a set of table tennis ball and golf ball images. The UniT encoder contains 75~M parameters.

\textbf{T3} \cite{zhao2024transferable} is built upon vision transformer backbone and is pre-trained on a large-scale dataset collected from multiple tactile sensors. The pretraining includes both masked autoencoding for reconstruction and supervision on perception-related downstream tasks. The resulting representation demonstrates strong generalization and broad applicability across diverse tasks and sensors. We use the T3-medium encoder released by T3 \cite{zhao2023learning} for both simulation and real-world experiments. The model contains 86~M parameters.

\textbf{AnyTouch} \cite{feng2025anytouch} is also a tactile representation built upon vision transformer backbone. It is pre-trained on dataset that includes aligned tactile images and videos collected from four different visuo-tactile sensors. AnyTouch jointly leverages pixel-level reconstruction and multi-sensor alignment objectives to learn semantic features. It enables robust perception and effective cross-sensor transfer, demonstrating strong generalization across both static and dynamic tasks. We use the checkpoint released by AnyTouch \cite{feng2025anytouch} for both simulation and real-world experiments. The model’s encoder is based on a ViT-Large backbone, with a total of 305~M parameters. As shown in Table~\ref{tab:training_time_comparison}, the large size of AnyTouch significantly slows down training for policies, reducing its flexibility.

\subsection{Policies}
\textbf{Diffusion Policy}~\cite{chi2023diffusionpolicy} formulates action generation as a conditional denoising diffusion process. During training, Gaussian noise is progressively added to ground-truth action sequences to construct a forward diffusion process. A denoising model is then trained to reverse this process by predicting noises, conditioned on visual observations. At inference time, the model  performs iterative denoising to sample action sequences, enabling robust and multimodal policy learning from demonstration data. We use CNN-based Diffusion Policy, where the denoising model is a 1D UNet \cite{chi2023diffusionpolicy}.

\textbf{Equivariant Diffusion Policy}~\cite{wang2024equivariant}
extends the diffusion framework by incorporating SO(2) equivariance into both the observation and action representations. This equivariant design enhances generalization and data efficiency, particularly in 6-DoF manipulation tasks exhibiting rotational symmetries. We adopt the image-based Equivariant Diffusion Policy, as all policies operate on RGB visual observations. Additionally, we use the relative control mode of EquiDiff, since both the simulation and real-world environments use a relative action space.

\textbf{Flow Matching}~\cite{zhang2024flowpolicy}
reformulates trajectory generation as learning a continuous flow field that maps random waypoints to actions in a single step. Unlike diffusion models, it avoids iterative denoising by solving an optimal transport-inspired regression problem. This one-shot inference process enables faster policy execution while maintaining comparable performance to diffusion-based methods in many robotic manipulation benchmarks.

\section{Additional Details on Experiments} \label{app:experiment_details}

\subsection{Experimental Setup}

Table~\ref{tab:task_setup} summarizes the task configurations used for policy training and evaluation in both simulation and real-world experiments, including the number of cameras, tactile sensors, and training demonstrations per task. Two simulation setups are used in this work. The first setup, corresponding to the results reported in Sections~\ref{sec:modality-benchmark} --~\ref{sec:tac-rep-benchmark}, is used to benchmark the proposed ManiFeel task suite and analyze the contribution of different sensing modalities, tactile representations, and policy architectures. The second simulation setup mirrors the number of demonstrations and conditions of the real-world experiments (Section~\ref{sec:real_world_eval}) to enable a direct comparison between simulated and physical environments. In all experiments, only tactile signals from the right finger sensor are used. For real-world experiments, we employ an OSC-yaw controller that regulates the end-effector position and yaw rotation while keeping roll and pitch fixed throughout execution.

\begin{table}[t]
\centering
\caption{\textbf{Tasks setup in the experiments.} Ctrl: control type; Cam: number of cameras; Tac: number of tactile sensors; ActD: action dimension; Demo: number of demonstrations used for training; Lighting: lighting condition; Occlusion: whether the camera view is occluded.}
\label{tab:task_setup}
\vspace{1mm}

\resizebox{\columnwidth}{!}{%
\begin{tabular}{lccccccc}
\toprule
\textbf{Task} & \textbf{Ctrl} & \textbf{Cam} & \textbf{Tac} & \textbf{ActD} & \textbf{Demo} & \textbf{Lighting} & \textbf{Occlusion} \\
\midrule
\multicolumn{8}{c}{Simulation Experiment (Sections~\ref{sec:modality-benchmark} --~\ref{sec:tac-rep-benchmark})} \\
\midrule
Peg Insertion               & Relative Pose & 1 (Wrist) & 1 & 6 & 50  & Normal & No \\
USB Insertion               & Relative Pose & 1 (Wrist) & 1 & 6 & 50  & Normal & No \\
Power Plug Insertion        & Relative Pose & 1 (Wrist) & 1 & 6 & 50  & Normal & No \\
Gear Assembly               & Relative Pose & 1 (Wrist) & 1 & 6 & 50  & Normal & No \\
Nut--bolt Assembly               & Relative Pose & 1 (Wrist) & 1 & 6 & 30  & Normal & No \\
Bulb Installation                & Relative Pose & 1 (Wrist) & 1 & 6 & 20  & Normal & No \\
Peg Reorientation           & Relative Pose & 1 (Front) & 1 & 6 & 50  & Normal & Yes \\
Object Search               & Relative Pose & 1 (Front) & 1 & 7 & 50  & Normal & Yes \\
Ball Sorting (Dim)          & Relative Pose & 1 (Front) & 1 & 7 & 50  & Dim    & No \\
\midrule
\multicolumn{8}{c}{Simulation Experiment (Section~\ref{sec:real_world_eval})} \\
\midrule
Gear Assembly               & Relative Pose & 1 (Wrist) & 1 & 6 & 80  & Normal & No \\
Bulb Installation           & Relative Pose & 1 (Wrist) & 1 & 5 & 20 & Normal & No \\
Ball Sorting (Normal)       & Relative Pose & 1 (Front) & 1 & 7 & 80 & Normal & No \\
Ball Sorting (Dim)          & Relative Pose & 1 (Front) & 1 & 7 & 80  & Dim    & No \\
\midrule
\multicolumn{8}{c}{Real-world Experiment (Section~\ref{sec:real_world_eval})} \\
\midrule
Gear Assembly               & Relative Pose & 1 (Wrist) & 1 & 5 & 80 & Normal & No \\
Bulb Installation           & Relative Pose & 1 (Wrist) & 1 & 5 & 20  & Normal    & No \\
Ball Sorting (Normal)       & Relative Pose & 1 (Front) & 1 & 5 & 80 & Normal & No \\
Ball Sorting (Dim)          & Relative Pose & 1 (Front) & 1 & 5 & 80  & Dim    & No \\
\bottomrule
\end{tabular}%
}
\end{table}

\begin{figure}[ht]
    \centering
    \includegraphics[width=0.78\linewidth]{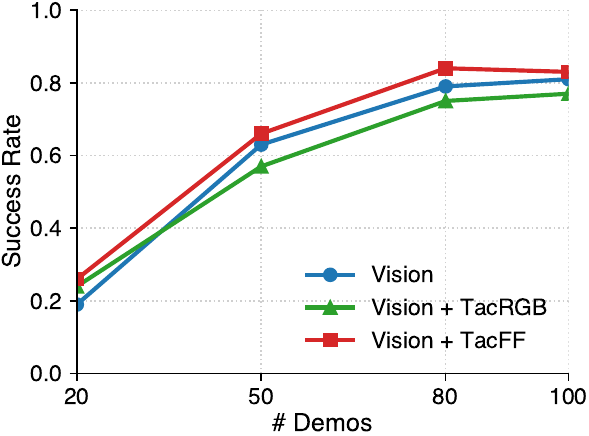}
    \caption{Success rate of gear assembly with increasing demonstrations across sensing configurations. Performance saturates around $80$ demos, indicating a balance point between data efficiency and policy performance.}
    \label{fig:demo_scaling}
\end{figure}

To determine an appropriate number of demonstrations for training, we conducted a scaling study on the gear assembly task, as illustrated in Fig.~\ref{fig:demo_scaling}. The results show that success rates increase steadily with the number of demonstrations but begin to saturate at around $80$ demonstrations across all sensing configurations. Therefore, we adopt $80$ demonstrations as the shared training set size for both simulation and real-world comparative experiments on gear assembly and ball sorting tasks, balancing data efficiency and policy stability. For the bulb installation task, we use $20$ demonstrations due to the inherently longer teleoperation trajectories, which result in substantially heavier datasets and higher training costs. Despite the smaller dataset, the policy achieves a high success rate (over $70\%$) in simulation, indicating that performance has already converged within this data regime. This choice thus represents a practical trade-off between data collection efficiency and policy performance.

All code, datasets, and configuration files used for training and rollout will be released publicly to facilitate full reproducibility and extension of the ManiFeel benchmark.

\subsection{Training Details} \label{app:training_details}
We summarize the training and inference hyperparameters in Table~\ref{tab:hyperparams}, which are consistent across all policies. The original visual image resolution is $256{\times}256$, while the tactile images and tactile force fields are recorded at $240{\times}320$ and $10{\times}14$, respectively. For training, we consistently resize tactile images to $256{\times}256$ when using the ResNet18 tactile encoder and UniT tactile representation, while visual inputs are maintained at $256{\times}256$ for the visual encoder. For T3 and AnyTouch, we use $224{\times}224$ tactile image inputs to match the ViT backbone architecture. For Diffusion Policy, we adopt DDPM for simulation experiments and switch to DDIM during real-world experiments to enable faster inference. For Equivariant Diffusion Policy (EquiDP), we use DDIM.

\begin{table}[t]
  \centering
  \caption{\textbf{Hyperparameters for training and inference.} 
  To: observation horizon; Ta: action horizon; Tp: action prediction horizon; 
  ImgRes: environment RGB image resolution; TacImgRes: tactile image resolution; 
  TacFFRes: tactile force-field resolution; 
  Lr: learning rate; DiffT.: number of training diffusion iterations; 
  DiffE.: number of inference diffusion iterations.}
  \label{tab:hyperparams}
  \vspace{1mm}

  \resizebox{\columnwidth}{!}{%
  \begin{tabular}{l|*{9}{c}}
    \toprule
    \textbf{H-Param} & \textbf{To} & \textbf{Ta} & \textbf{Tp} & \textbf{ImgRes} & \textbf{TacImgRes} & \textbf{TacFFRes} & \textbf{Lr} & \textbf{DiffT} & \textbf{DiffE} \\
    \midrule
    \multicolumn{10}{c}{Simulation} \\
    \midrule
    Peg Insertion           & 2 & 8 & 16 & 256×256 & 240×320 & 10×14 & 8e-5 & 100 & 100 \\
    USB Insertion           & 2 & 8 & 16 & 256×256 & 240×320 & 10×14 & 8e-5 & 100 & 100 \\
    Power Plug Insertion    & 2 & 8 & 16 & 256×256 & 240×320 & 10×14 & 8e-5 & 100 & 100 \\
    Gear Assembly           & 2 & 8 & 16 & 256×256 & 240×320 & 10×14 & 8e-5 & 100 & 100 \\
    Nut--bolt Assembly      & 2 & 8 & 16 & 256×256 & 240×320 & 10×14 & 8e-5 & 100 & 100 \\
    Bulb Installation       & 2 & 8 & 16 & 256×256 & 240×320 & 10×14 & 8e-5 & 100 & 100 \\
    Peg Reorientation       & 2 & 8 & 16 & 256×256 & 240×320 & 10×14 & 8e-5 & 100 & 100 \\
    Object Search           & 2 & 8 & 16 & 256×256 & 240×320 & 10×14 & 8e-5 & 100 & 100 \\
    Ball Sorting            & 2 & 8 & 16 & 256×256 & 240×320 & 10×14 & 8e-5 & 100 & 100 \\
    \midrule
    \multicolumn{10}{c}{Real-World} \\
    \midrule
    Gear Assembly           & 2 & 6 & 16 & 320×240 & 160×240 & 10×14 & 8e-5 & 75  & 16  \\
    Bulb Installation       & 2 & 8 & 16 & 320×240 & 160×240 & 10×14 & 8e-5 & 75  & 16  \\
    Ball Sorting (Normal)   & 2 & 6 & 16 & 320×240 & 160×240 & 10×14 & 8e-5 & 75  & 16  \\
    Ball Sorting (Dim)      & 2 & 8 & 16 & 320×240 & 160×240 & 10×14 & 8e-5 & 75  & 16  \\
    \bottomrule
  \end{tabular}%
  }
\end{table}



\subsection{Runtime Analysis of Tactile Representations}
In this section, we evaluate the training time required for different tactile encoders and representations, including ResNet18, UniT, T3, and AnyTouch. Table~\ref{tab:training_time_comparison} summarizes the training time per epoch and total training time (over 20 epochs) for visuotactile policies (Vision+TacRGB sensing configuration) in the ball sorting task. All experiments are conducted using an NVIDIA A30 GPU with a batch size of 8. AnyTouch incurs substantially higher computational cost, with each training epoch taking over 67 minutes. In contrast, UniT and T3 are more efficient, with training times of approximately 26 and 18 minutes per epoch, respectively, making them better suited for faster policy iteration and development.

\begin{table}[ht]
\centering
\small
\caption{Average training time per epoch and total training time (over 20 epochs) for visuotactile policies with different tactile representations in the real-world sorting task under normal lighting conditions. The dataset includes 103 human demonstrations. Time per epoch is reported in minutes and total time in hours.}
\label{tab:training_time_comparison}
\begin{tabular}{lcccc}
\toprule
\textbf{Metric} & \textbf{ResNet18} & \textbf{UniT} & \textbf{T3} & \textbf{AnyTouch} \\
\midrule
Time per Epoch (min) & $8.1$   & $25.8$  & $17.7$  & $66.3$  \\
Total Time (hr)      & $2.7$   & $8.6$   & $5.9$   & $22.1$  \\
\bottomrule
\end{tabular}
\end{table}

\subsection{Experiment on Different Camera Viewpoints}
\begin{figure}[ht]
    \centering

    \subfloat[Observations\label{fig:tactile_explore_obs}]{%
        \includegraphics[width=0.30\linewidth]{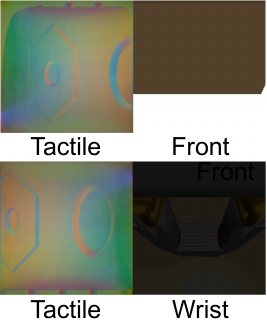}}
    \quad\quad\quad
    \subfloat[Success rate\label{fig:tactile_explore_barplot}]{%
        \includegraphics[width=0.45\linewidth]{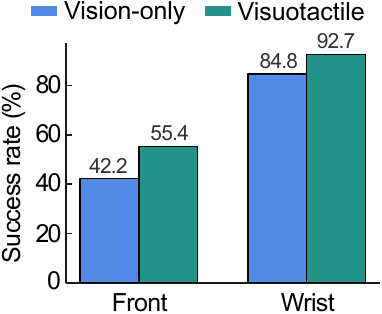}}
    \caption{Comparison of front and wrist camera views for vision-only and visuotactile (vision + tacRGB) policies in the object-search task. (a) Sample observations showing tactile inputs and visual images from occluded front views and dim wrist views. (b) Final success rates (\%) averaged over the last ten epochs and three seeds.}
    \label{fig:usb_viewpoint_comparison}
\end{figure}

Fig.~\ref{fig:usb_viewpoint_comparison} illustrates the performance of policies in the object-search task when using visual observations from two different camera viewpoints, specifically wrist-mounted and front-facing perspectives. Fig.~\ref{fig:tactile_explore_obs} illustrates tactile inputs and visual images from occluded front views and dim wrist views. The result shows that both vision-only and visuotactile (vision+tacRGB) policies achieve higher success rates when trained with wrist-mounted inputs (see Fig.~\ref{fig:tactile_explore_barplot}). Moreover, the visuotactile policy consistently outperforms its vision-only counterpart under both viewpoints, confirming that incorporating tactile sensing enhances performance under visually limited conditions. 

\bibliographystyle{IEEEtran}
\bibliography{ref}

@inproceedings{he2016deep,
  title={Deep residual learning for image recognition},
  author={He, Kaiming and Zhang, Xiangyu and Ren, Shaoqing and Sun, Jian},
  booktitle={ Proc. IEEE Conf. Comput. Vis. Pattern Recognit.},
  year={2016}
}

@article{she2021cable,
  title={Cable manipulation with a tactile-reactive gripper},
  author={She, Yu and Wang, Shaoxiong and Dong, Siyuan and Sunil, Neha and Rodriguez, Alberto and Adelson, Edward},
  journal={Int. J. Robot. Res.},
  volume={40},
  number={12-14},
  pages={1385--1401},
  year={2021},
}

@inproceedings{hogan2020tactile,
  title={{Tactile Dexterity}: Manipulation primitives with tactile feedback},
  author={Hogan, Francois R and Ballester, Jose and Dong, Siyuan and Rodriguez, Alberto},
  booktitle={Proc. IEEE Int. Conf. Robot. Autom.},
  pages={8863--8869},
  year={2020}
}

@inproceedings{sunil2022visuotactile,
  title={Visuotactile Affordances for Cloth Manipulation with Local Control},
  author={Sunil, Neha and Wang, Shaoxiong and She, Yu and Adelson, Edward and Garcia, Alberto Rodriguez},
  booktitle={Proc. Conf. Robot Learn.},
  year={2022}
}

@inproceedings{kim2021active,
  title={Active extrinsic contact sensing: Application to general peg-in-hole insertion},
  author={Kim, Sangwoon and Rodriguez, Alberto},
  booktitle={Proc. IEEE Int. Conf. Robot. Autom.},
  pages={10241--10247},
  year={2022}
}

@inproceedings{tian2019manipulation,
  title={{Manipulation by Feel}: Touch-based control with deep predictive models},
  author={Tian, Stephen and Ebert, Frederik and Jayaraman, Dinesh and Mudigonda, Mayur and Finn, Chelsea and Calandra, Roberto and Levine, Sergey},
  booktitle={Proc. IEEE Int. Conf. Robot. Autom.},
  pages={818--824},
  year={2019}
}

@inproceedings{calandra2017feeling,
  title={{The Feeling of Success}: Does Touch Sensing Help Predict Grasp Outcomes?},
  author={Calandra, Roberto and Owens, Andrew and Upadhyaya, Manu and Yuan, Wenzhen and Lin, Justin and Adelson, Edward H and Levine, Sergey},
  booktitle={Proc. Conf. Robot Learn.},
  pages={314--323},
  year={2017}
}

@inproceedings{si2022grasp,
  title={Grasp Stability Prediction with Sim-to-Real Transfer from Tactile Sensing},
  author={Si, Zilin and Zhu, Zirui and Agarwal, Arpit and Anderson, Stuart and Yuan, Wenzhen},
  booktitle={Proc. IEEE/RSJ Int. Conf. Intell. Robots Syst.},
  pages={7809--7816},
  year={2022}
}

@inproceedings{kanitkar2022poseit,
  title={{PoseIt}: A Visual-Tactile Dataset of Holding Poses for Grasp Stability Analysis},
  author={Kanitkar, Shubham and Jiang, Helen and Yuan, Wenzhen},
  booktitle={Proc. IEEE/RSJ Int. Conf. Intell. Robots Syst.},
  pages={71--78},
  year={2022}
}

@article{calandra2018more,
  title={{More Than a Feeling}: Learning to grasp and regrasp using vision and touch},
  author={Calandra, Roberto and Owens, Andrew and Jayaraman, Dinesh and Lin, Justin and Yuan, Wenzhen and Malik, Jitendra and Adelson, Edward H and Levine, Sergey},
  journal={IEEE Robotics and Automation Letters},
  volume={3},
  number={4},
  pages={3300--3307},
  year={2018},
}

@inproceedings{hogan2018tactile,
  title={{Tactile Regrasp}: Grasp adjustments via simulated tactile transformations},
  author={Hogan, Francois R and Bauza, Maria and Canal, Oleguer and Donlon, Elliott and Rodriguez, Alberto},
  booktitle={Proc. IEEE/RSJ Int. Conf. Intell. Robots Syst.},
  pages={2963--2970},
  year={2018}
}

@article{han2021learning,
  title={Learning Generalizable Vision-Tactile Robotic Grasping Strategy for Deformable Objects via Transformer},
  author={Han, Yunhai and Batra, Rahul and Boyd, Nathan and Zhao, Tuo and She, Yu and Hutchinson, Seth and Zhao, Ye},
  journal={arXiv preprint arXiv:2112.06374},
  year={2021}
}

@inproceedings{feng2020center,
  title={Center-of-mass-based robust grasp planning for unknown objects using tactile-visual sensors},
  author={Feng, Qian and Chen, Zhaopeng and Deng, Jun and Gao, Chunhui and Zhang, Jianwei and Knoll, Alois},
  booktitle={Proc. IEEE Int. Conf. Robot. Autom.},
  pages={610--617},
  year={2020}
}

@article{yuan2017gelsight,
  title={{GelSight}: High-resolution robot tactile sensors for estimating geometry and force},
  author={Yuan, Wenzhen and Dong, Siyuan and Adelson, Edward H},
  journal={Sensors},
  year={2017},
}

@article{lloyd2021goal,
  title={Goal-driven robotic pushing using tactile and proprioceptive feedback},
  author={Lloyd, John and Lepora, Nathan F},
  journal={IEEE Transactions on Robotics},
  volume={38},
  number={2},
  pages={1201--1212},
  year={2021},
}

@article{zheng2022autonomous,
  title={Autonomous learning of page flipping movements via tactile feedback},
  author={Zheng, Yi and Veiga, Filipe Fernandes and Peters, Jan and Santos, Veronica J},
  journal={IEEE Transactions on Robotics},
  volume={38},
  number={5},
  pages={2734--2749},
  year={2022},
}

@article{shirai2023tactile,
  title={Tactile Tool Manipulation},
  author={Shirai, Yuki and Jha, Devesh K and Raghunathan, Arvind U and Hong, Dennis},
  journal={arXiv preprint arXiv:2301.06698},
  year={2023}
}

@inproceedings{wang2021gelsight,
  title={{GelSight Wedge}: Measuring high-resolution {3D} contact geometry with a compact robot finger},
  author={Wang, Shaoxiong and She, Yu and Romero, Branden and Adelson, Edward},
  booktitle={Proc. IEEE Int. Conf. Robot. Autom.},
  pages={6468--6475},
  year={2021}
}

@inproceedings{mandlekar2022matters,
  title={What Matters in Learning from Offline Human Demonstrations for Robot Manipulation},
  author={Mandlekar, Ajay and Xu, Danfei and Wong, Josiah and Nasiriany, Soroush and Wang, Chen and Kulkarni, Rohun and Fei-Fei, Li and Savarese, Silvio and Zhu, Yuke and Mart{\'\i}n-Mart{\'\i}n, Roberto},
  booktitle={Proc. Conf. Robot Learn.},
  pages={1678--1690},
  year={2022},
}

@inproceedings{zhao2023learning,
  title={Learning fine-grained bimanual manipulation with low-cost hardware},
  author={Zhao, Tony Z and Kumar, Vikash and Levine, Sergey and Finn, Chelsea},
booktitle={Proceedings of Robotics: Science and Systems},
  year={2023}
}

@inproceedings{chi2023diffusionpolicy,
	title={Diffusion Policy: Visuomotor Policy Learning via Action Diffusion},
	author={Chi, Cheng and Feng, Siyuan and Du, Yilun and Xu, Zhenjia and Cousineau, Eric and Burchfiel, Benjamin and Song, Shuran},
	booktitle={Proceedings of Robotics: Science and Systems},
	year={2023}
}

@article{lee2024behavior,
  title={Behavior Generation with Latent Actions},
  author={Lee, Seungjae and Wang, Yibin and Etukuru, Haritheja and Kim, H Jin and Shafiullah, Nur Muhammad Mahi and Pinto, Lerrel},
  journal={arXiv preprint arXiv:2403.03181},
  year={2024}
}

@article{chi2024universal,
  title={Universal Manipulation Interface: In-The-Wild Robot Teaching Without In-The-Wild Robots},
  author={Chi, Cheng and Xu, Zhenjia and Pan, Chuer and Cousineau, Eric and Burchfiel, Benjamin and Feng, Siyuan and Tedrake, Russ and Song, Shuran},
  journal={arXiv preprint arXiv:2402.10329},
  year={2024}
}

@article{fu2024mobile,
  title={Mobile aloha: Learning bimanual mobile manipulation with low-cost whole-body teleoperation},
  author={Fu, Zipeng and Zhao, Tony Z and Finn, Chelsea},
  journal={arXiv preprint arXiv:2401.02117},
  year={2024}
}

@article{wang2024dexcap,
  title={DexCap: Scalable and Portable Mocap Data Collection System for Dexterous Manipulation},
  author={Wang, Chen and Shi, Haochen and Wang, Weizhuo and Zhang, Ruohan and Fei-Fei, Li and Liu, C Karen},
  journal={arXiv preprint arXiv:2403.07788},
  year={2024}
}

@misc{team2023octo,
  title={Octo: An open-source generalist robot policy},
  author={Team, Octo Model and Ghosh, Dibya and Walke, Homer and Pertsch, Karl and Black, Kevin and Mees, Oier and Dasari, Sudeep and Hejna, Joey and Xu, Charles and Luo, Jianlan and others},
  year={2023}
}

@inproceedings{ze20243d,
  title={{3D} diffusion policy},
  author={Ze, Yanjie and Zhang, Gu and Zhang, Kangning and Hu, Chenyuan and Wang, Muhan and Xu, Huazhe},
booktitle={Proceedings of Robotics: Science and Systems},
  year={2024}
}

@ARTICLE{xu2024letac,
  author={Xu, Zhengtong and She, Yu},
  journal={IEEE Transactions on Robotics}, 
  title={{LeTac-MPC}: Learning Model Predictive Control for Tactile-Reactive Grasping}, 
  year={2024},
  doi={10.1109/TRO.2024.3463470}}

@inproceedings{wang2024poco,
  title={PoCo: Policy Composition from and for Heterogeneous Robot Learning},
  author={Wang, Lirui and Zhao, Jialiang and Du, Yilun and Adelson, Edward H and Tedrake, Russ},
booktitle={Proceedings of Robotics: Science and Systems},
  year={2024}
}

@inproceedings{guzey2023see,
  title={See to touch: Learning tactile dexterity through visual incentives},
  author={Guzey, Irmak and Dai, Yinlong and Evans, Ben and Chintala, Soumith and Pinto, Lerrel},
  booktitle={Proceedings of International Conference on Robotics and Automation},
  year={2024}
}

@inproceedings{yang2023seq2seq,
  title={Seq2Seq Imitation Learning for Tactile Feedback-based Manipulation},
  author={Yang, Wenyan and Angleraud, Alexandre and Pieters, Roel S and Pajarinen, Joni and K{\"a}m{\"a}r{\"a}inen, Joni-Kristian},
  booktitle={Proceedings of IEEE International Conference on Robotics and Automation},
  pages={5829--5836},
  year={2023},
}

@inproceedings{yu2023mimictouch,
  title={MimicTouch: Leveraging Multi-modal Human Tactile Demonstrations for Contact-rich Manipulation},
  author={Yu, Kelin and Han, Yunhai and Wang, Qixian and Saxena, Vaibhav and Xu, Danfei and Zhao, Ye},
  booktitle={Proceedings of Conference on Robot Learning},
  year={2024}
}

@article{liu2024maniwav,
    title={ManiWAV: Learning Robot Manipulation from In-the-Wild Audio-Visual Data},
    author={Liu, Zeyi and Chi, Cheng and Cousineau, Eric and Kuppuswamy, Naveen and Burchfiel, Benjamin and Song, Shuran},
    journal={arXiv preprint arXiv:2406.19464},
    year={2024}
}

@inproceedings{lin2024learning,
  author={Lin, Toru and Zhang, Yu and Li, Qiyang and Qi, Haozhi and Yi, Brent and Levine, Sergey and Malik, Jitendra},
  title={Learning Visuotactile Skills with Two Multifingered Hands},
  booktitle={Proceedings of International Conference on Robotics and Automation},
  year={2025}
}

@inproceedings{zhao2024transferable,
    title={Transferable Tactile Transformers for Representation Learning Across Diverse Sensors and Tasks}, 
    author={Jialiang Zhao and Yuxiang Ma and Lirui Wang and Edward H. Adelson},
    year={2024},
  booktitle={Proceedings of Conference on Robot Learning},
}

@inproceedings{pari2021surprising,
  title={The surprising effectiveness of representation learning for visual imitation},
  author={Pari, Jyothish and Shafiullah, Nur Muhammad and Arunachalam, Sridhar Pandian and Pinto, Lerrel},
booktitle={Proceedings of Robotics: Science and Systems},
  year={2022}
}

@article{polic2019convolutional,
  title={Convolutional autoencoder for feature extraction in tactile sensing},
  author={Polic, Marsela and Krajacic, Ivona and Lepora, Nathan and Orsag, Matko},
  journal={IEEE Robotics and Automation Letters},
  volume={4},
  number={4},
  pages={3671--3678},
  year={2019},
}

@inproceedings{cao2023learn,
  title={Learn from Incomplete Tactile Data: Tactile Representation Learning with Masked Autoencoders},
  author={Cao, Guanqun and Jiang, Jiaqi and Bollegala, Danushka and Luo, Shan},
  booktitle={Proceedings of IEEE/RSJ International Conference on Intelligent Robots and Systems},
  pages={10800--10805},
  year={2023},
}

@inproceedings{sferrazza2023power,
  title={The power of the senses: Generalizable manipulation from vision and touch through masked multimodal learning},
  author={Sferrazza, Carmelo and Seo, Younggyo and Liu, Hao and Lee, Youngwoon and Abbeel, Pieter},
  booktitle={Proceedings of IEEE/RSJ International Conference on Intelligent Robots and Systems},
  year={2024},
}

@inproceedings{yang2024unitouch,
  title={Binding touch to everything: Learning unified multimodal tactile representations},
  author={Yang, Fengyu and others},
  booktitle={Proceedings of the IEEE/CVF Conference on Computer Vision and Pattern Recognition},
  pages={26340--26353},
  year={2024}
}

@inproceedings{higuera2024sparsh,
  title={Sparsh: Self-supervised touch representations for vision-based tactile sensing},
  author={Higuera, Carolina and Sharma, Akash and others},
  booktitle={Proceedings of Conference on Robot Learning},
  year={2024}
}

@misc{xu2025unit,
      title={{UniT}: Data Efficient Tactile Representation with Generalization to Unseen Objects}, 
      author={Zhengtong Xu and Raghava Uppuluri and Xinwei Zhang and Cael Fitch and Philip Glen Crandall and Wan Shou and Dongyi Wang and Yu She},
      year={2025},
      eprint={2408.06481},
      archivePrefix={arXiv},
      primaryClass={cs.RO},
      url={https://arxiv.org/abs/2408.06481}, 
}

@article{hou2024adaptive,
  title={Adaptive Compliance Policy: Learning Approximate Compliance for Diffusion Guided Control},
  author={Hou, Yifan and Liu, Zeyi and Chi, Cheng and Cousineau, Eric and Kuppuswamy, Naveen and Feng, Siyuan and Burchfiel, Benjamin and Song, Shuran},
  journal={arXiv preprint arXiv:2410.09309},
  year={2024}
}

@article{bhirangi2024anyskin,
  title={Anyskin: Plug-and-play skin sensing for robotic touch},
  author={Bhirangi, Raunaq and Pattabiraman, Venkatesh and Erciyes, Enes and Cao, Yifeng and Hellebrekers, Tess and Pinto, Lerrel},
  journal={arXiv preprint arXiv:2409.08276},
  year={2024}
}

@article{xue2025reactive,
  title     = {Reactive Diffusion Policy: Slow-Fast Visual-Tactile Policy Learning for Contact-Rich Manipulation},
  author    = {Xue, Han and Ren, Jieji and Chen, Wendi and Zhang, Gu and Fang, Yuan and Gu, Guoying and Xu, Huazhe and Lu, Cewu},
  journal   = {arXiv preprint arXiv:2503.02881},
  year      = {2025}
}

@inproceedings{
huang2024dvitac,
title={3D-ViTac: Learning Fine-Grained Manipulation with Visuo-Tactile Sensing},
author={Binghao Huang and Yixuan Wang and Xinyi Yang and Yiyue Luo and Yunzhu Li},
booktitle={Proceedings of Conference on Robot Learning},
year={2024},
url={https://openreview.net/forum?id=bk28WlkqZn}
}

@article{xue2025demogen,
  title={Demogen: Synthetic demonstration generation for data-efficient visuomotor policy learning},
  author={Xue, Zhengrong and Deng, Shuying and Chen, Zhenyang and Wang, Yixuan and Yuan, Zhecheng and Xu, Huazhe},
  journal={arXiv preprint arXiv:2502.16932},
  year={2025}
}

@article{zhang2024affordance,
  title={Affordance-based Robot Manipulation with Flow Matching},
  author={Zhang, Fan and Gienger, Michael},
  journal={arXiv preprint arXiv:2409.01083},
  year={2024}
}

@article{zhao2024aloha,
  title={Aloha unleashed: A simple recipe for robot dexterity},
  author={Zhao, Tony Z and Tompson, Jonathan and Driess, Danny and Florence, Pete and Ghasemipour, Kamyar and Finn, Chelsea and Wahid, Ayzaan},
  journal={arXiv preprint arXiv:2410.13126},
  year={2024}
}

@article{black2410pi0,
  title={$\pi$0: A vision-language-action flow model for general robot control, 2024},
  author={Black, Kevin and Brown, Noah and Driess, Danny and Esmail, Adnan and Equi, Michael and Finn, Chelsea and Fusai, Niccolo and Groom, Lachy and Hausman, Karol and Ichter, Brian and others},
  journal={URL https://arxiv. org/abs/2410.24164},
year = {2024}
}

@article{zhao2024tac,
  title={Tac-Man: Tactile-informed prior-free manipulation of articulated objects},
  author={Zhao, Zihang and Li, Yuyang and Li, Wanlin and Qi, Zhenghao and Ruan, Lecheng and Zhu, Yixin and Althoefer, Kaspar},
  journal={IEEE Transactions on Robotics},
  year={2024},
}

@article{gupta2025sensor,
  title={Sensor-Invariant Tactile Representation},
  author={Gupta, Harsh and Mo, Yuchen and Jin, Shengmiao and Yuan, Wenzhen},
  journal={arXiv preprint arXiv:2502.19638},
  year={2025}
}

@article{feng2025anytouch,
  title={AnyTouch: Learning Unified Static-Dynamic Representation across Multiple Visuo-tactile Sensors},
  author={Feng, Ruoxuan and Hu, Jiangyu and Xia, Wenke and Gao, Tianci and Shen, Ao and Sun, Yuhao and Fang, Bin and Hu, Di},
  journal={arXiv preprint arXiv:2502.12191},
  year={2025}
}

@inproceedings{
  wang2024equivariant,
  title={Equivariant Diffusion Policy},
  author={Dian Wang and Stephen Hart and David Surovik and Tarik Kelestemur and Haojie Huang and Haibo Zhao and Mark Yeatman and Jiuguang Wang and Robin Walters and Robert Platt},
  booktitle={8th Annual Conference on Robot Learning},
  year={2024},
  url={}
}

@ARTICLE{10947008,
  author={Aslam, Shoaib and Kumar, Krish and Zhou, Pokuang and Yu, Hongyu and Yu Wang, Michael and She, Yu},
  journal={IEEE Transactions on Automation Science and Engineering}, 
  title={DartBot: Overhand Throwing of Deformable Objects With Tactile Sensing and Reinforcement Learning}, 
  year={2025},
  volume={22},
  pages={13644-13661},
  doi={10.1109/TASE.2025.3556875}}

@inproceedings{du2024stick,
  title={Stick Roller: A Case Study on Efficiently Learning Precise Tactile Dynamics for In-hand Manipulation},
  author={Du, Yipai and Zhou, Pokuang and Wang, Michael Yu and Lian, Wenzhao and She, Yu},
  booktitle={Proceedings of IEEE/RSJ International Conference on Intelligent Robots and Systems},
  year={2024},
  pages={2312--2318},
}

@article{mandlekar2023mimicgen,
  title={Mimicgen: A data generation system for scalable robot learning using human demonstrations},
  author={Mandlekar, Ajay and Nasiriany, Soroush and Wen, Bowen and Akinola, Iretiayo and Narang, Yashraj and Fan, Linxi and Zhu, Yuke and Fox, Dieter},
  journal={arXiv preprint arXiv:2310.17596},
  year={2023}
}

@article{tao2024maniskill3,
  title={Maniskill3: Gpu parallelized robotics simulation and rendering for generalizable embodied ai},
  author={Tao, Stone and Xiang, Fanbo and Shukla, Arth and Qin, Yuzhe and Hinrichsen, Xander and Yuan, Xiaodi and Bao, Chen and Lin, Xinsong and Liu, Yulin and Chan, Tse-kai and others},
  journal={arXiv preprint arXiv:2410.00425},
  year={2024}
}

@ARTICLE{akinola2024tacsl,
  author={Akinola, Iretiayo and Xu, Jie and Carius, Jan and Fox, Dieter and Narang, Yashraj},
  journal={IEEE Transactions on Robotics}, 
  title={TacSL: A Library for Visuotactile Sensor Simulation and Learning}, 
  year={2025},
  volume={41},
  pages={2645-2661},
  doi={10.1109/TRO.2025.3547267}}

@misc{IsaacGym,
      title={Isaac Gym: High Performance GPU-Based Physics Simulation For Robot Learning}, 
      author={Viktor Makoviychuk and Lukasz Wawrzyniak and Yunrong Guo and Michelle Lu and Kier Storey and Miles Macklin and David Hoeller and Nikita Rudin and Arthur Allshire and Ankur Handa and Gavriel State},
      year={2021},
      eprint={2108.10470},
      archivePrefix={arXiv},
      primaryClass={cs.RO},
      url={https://arxiv.org/abs/2108.10470}, 
}

@ARTICLE{quan2023simtacls,
  author={Luu, Quan Khanh and Nguyen, Nhan Huu and Ho, Van Anh},
  journal={IEEE Transactions on Robotics}, 
  title={Simulation, Learning, and Application of Vision-Based Tactile Sensing at Large Scale}, 
  year={2023},
  volume={39},
  number={3},
  pages={2003-2019},
  doi={10.1109/TRO.2023.3245983}}

@ARTICLE{quan2025protac,
  author={Luu, Quan Khanh and Nguyen, Dinh Quang and Nguyen, Nhan Huu and Dam, Nam Phuong and Ho, Van Anh},
  journal={IEEE Transactions on Robotics}, 
  title={Vision-Based Proximity and Tactile Sensing for Robot Arms: Design, Perception, and Control}, 
  year={2025},
  volume={41},
  number={},
  pages={5000-5019},
  doi={10.1109/TRO.2025.3593087}}

@misc{zhang2024flowpolicy,
      title={FlowPolicy: Enabling Fast and Robust 3D Flow-based Policy via Consistency Flow Matching for Robot Manipulation}, 
      author={Qinglun Zhang and Zhen Liu and Haoqiang Fan and Guanghui Liu and Bing Zeng and Shuaicheng Liu},
      year={2024},
      eprint={2412.04987},
      archivePrefix={arXiv},
      primaryClass={cs.RO},
      url={https://arxiv.org/abs/2412.04987}, 
}

@inproceedings{heusel2017fid,
  title={GANs trained by a two time-scale update rule converge to a local Nash equilibrium},
  author={Heusel, Martin and Ramsauer, Hubert and Unterthiner, Thomas and Nessler, Bernhard and Hochreiter, Sepp},
  booktitle={Advances in Neural Information Processing Systems},
  year={2017}
}

@inproceedings{binkowski2018kid,
  title={Demystifying MMD GANs},
  author={Bi{\'n}kowski, Miko{\l}aj and Sutherland, Dougal J. and Arbel, Michael and Gretton, Arthur},
  booktitle={International Conference on Learning Representations},
  year={2018}
}

@INPROCEEDINGS{zhu2017cyclegan,
  author={Zhu, Jun-Yan and Park, Taesung and Isola, Phillip and Efros, Alexei A.},
  booktitle={2017 IEEE International Conference on Computer Vision (ICCV)}, 
  title={Unpaired Image-to-Image Translation Using Cycle-Consistent Adversarial Networks}, 
  year={2017},
  volume={},
  number={},
  pages={2242-2251},
  doi={10.1109/ICCV.2017.244}}

@article{huang2025vtrefine,
title={VT-Refine: Learning Bimanual Assembly with Visuo-Tactile Feedback via Simulation Fine-Tuning},
author={Huang, Binghao and Xu, Jie and Akinola, Iretiayo and Yang, Wei and Sundaralingam, Balakumar and O'Flaherty, Rowland and Fox, Dieter and Wang, Xiaolong and Mousavian, Arsalan and Chao, Yu-Wei and Li, Yunzhu},
journal={9th Conference on Robot Learning},
year={2025}
}

@article{zhu2025touch,
    title={Touch in the wild: Learning fine-grained manipulation with a portable visuo-tactile gripper},
    author={Zhu, Xinyue and Huang, Binghao and Li, Yunzhu},
    journal={arXiv preprint arXiv:2507.15062},
    year={2025}
  }

@article{barreiros2025careful,
  title={A careful examination of large behavior models for multitask dexterous manipulation},
  author={Barreiros, Jose and Beaulieu, Andrew and Bhat, Aditya and Cory, Rick and Cousineau, Eric and Dai, Hongkai and Fang, Ching-Hsin and Hashimoto, Kunimatsu and Irshad, Muhammad Zubair and Itkina, Masha and others},
  journal={arXiv preprint arXiv:2507.05331},
  year={2025}
}

@article{zhou2025hand,
  title={In-Hand Singulation, Scooping, and Cable Untangling with a 5-Dof Tactile-Reactive Gripper},
  author={Zhou, Yuhao and Zhou, Pokuang and Wang, Shaoxiong and She, Yu},
  journal={Advanced Robotics Research},
  pages={202500020},
  year={2025},
  publisher={Wiley Online Library}
}

\end{document}